%% file: BugNIST_arXiv.tex
\documentclass[runningheads]{llncs}

\usepackage{eccv}

\usepackage{eccvabbrv}

\usepackage{graphicx}
\usepackage{booktabs}

\usepackage[accsupp]{axessibility}  

\usepackage[pagebackref,breaklinks,colorlinks,citecolor=eccvblue]{hyperref}

\usepackage{orcidlink}
\usepackage{booktabs}
\usepackage{tabularx}
\usepackage{tikz}

\newcommand{\bugnist}{BugNIST\xspace}

\begin{document}

\title{\bugnist\ -- a Large Volumetric Dataset for Object Detection under Domain Shift} 

\titlerunning{\bugnist\ -- Object Detection under Domain Shift}

\author{Patrick Møller Jensen\orcidlink{0000-0002-8479-4885} \and
Vedrana Andersen Dahl\orcidlink{0000-0001-6734-5570} \and
Rebecca Engberg\orcidlink{0009-0005-9546-1954} \and
Carsten Gundlach\orcidlink{0000-0002-2895-1882} \and Hans Marin Kjer\orcidlink{0000-0001-7900-5733} \and
Anders Bjorholm Dahl\orcidlink{0000-0002-0068-8170}}

\authorrunning{P.M.~Jensen et al.}

\institute{Technical University of Denmark, Kgs. Lyngby, Denmark
\email{\{patmjen,vand,reen\}@dtu.dk,cagu@fysik.dtu.dk,\{hmjk,abda\}@dtu.dk}}

\maketitle

\begin{abstract}
  Domain shift significantly influences the performance of deep learning algorithms, particularly for object detection within volumetric 3D images. Annotated training data is essential for deep learning-based object detection. However, annotating densely packed objects is time-consuming and costly. Instead, we suggest training models on individually scanned objects, causing a domain shift between training and detection data. To address this challenge, we introduce the \bugnist dataset, comprising 9154 micro-CT volumes of 12 bug types and 388 volumes of tightly packed bug mixtures. This dataset is characterized by having objects with the same appearance in the source and target domains, which is uncommon for other benchmark datasets for domain shift. During training, individual bug volumes labeled by class are utilized, while testing employs mixtures with center point annotations and bug type labels. Together with the dataset, we provide a baseline detection analysis, with the aim of advancing the field of 3D object detection methods. 
\end{abstract}

\definecolor{Background}{RGB}{83, 135, 221}
\definecolor{SL}{RGB}{218, 76, 76}
\definecolor{BC}{RGB}{71, 154, 95}
\definecolor{MA}{RGB}{125, 84, 178}
\definecolor{GH}{RGB}{232, 123, 159}
\definecolor{AC}{RGB}{229, 116, 57}
\definecolor{BP}{RGB}{135, 206, 191}
\definecolor{BF}{RGB}{197, 101, 199}
\definecolor{CF}{RGB}{237, 183, 50}
\definecolor{BL}{RGB}{91, 197, 219}
\definecolor{ML}{RGB}{34, 148, 135}
\definecolor{WO}{RGB}{240, 184, 153}
\definecolor{PP}{RGB}{160, 199, 92}

\begin{figure*}
    \input{images/illustration1_v2}
    \caption{
    Overview of the two parts of the dataset. In (a), representative individual bug volumes are displayed with a class abbreviation, name, and number of volumes. In (b), a few volumes of bug mixtures are shown. Each bug is annotated with bug class and center point.\label{fig:overview}}
\end{figure*}%

\section{Introduction}
\label{sec:intro}
Our work on domain shift in volumetric 3D images is motivated by the need for labeled data to train supervised deep learning models for volumetric imaging. The problem that we aim to solve is object detection and classification. We propose to label images of objects scanned as isolated entities as the source domain and use these as a basis to train models for object detection and classification in a complex context of mixed objects and other materials as the target domain. The effort needed for obtaining labels in the two domains is significantly different. If objects are isolated, they can be automatically labeled whereas mixed objects require expensive manual labeling. Automating the labeling based on isolated objects, however, leads to a domain shift between the data in the source domain for training and the data in the target domain for detection and classification. This domain shift is special because the appearance of the objects is the same in the two domains, but the surrounding context is different. We use the term \emph{context shift} for this type of domain shift. 

To promote the development of new deep learning-based methods that can handle context shift, we contribute the volumetric dataset called \bugnist\footnote{\url{https://abdahl.github.io/bugnist/}}. The dataset contains 9154 micro-CT volumes with individual bugs and 388 micro-CT volumes containing a mixture of several bugs as well as other materials such as leaves, wood shavings, \etc; see \cref{fig:overview}. The challenge is to use the individually scanned bugs to detect and classify the bugs in the mixtures. The dataset is intended to be a relatively simple test case for starting to develop methods for detecting objects in volumetric images under context shift, and bugs are chosen due to their complex shape and \emph{not} for solving a problem in entomology. 

As a basis for training an accurate deep learning model, a training dataset for object detection and classification must capture the appearance distribution of the object classes that are in the dataset. Since the appearance of a CT-scanned object changes minimally if it is scanned individually or as part of a mixture, the appearance distribution can be covered by only considering the individual scans. This is different from 2D photos. Here additional variations in scale, viewpoint, and illumination influence the appearance of objects, and it would be more difficult to acquire representative images of individual objects for a dataset based on 2D photos. 

Since \bugnist is a test case for context shift, we wanted to ensure that each class is covered well. By scanning systematically arranged bugs, we have generated 9154 volumes of individual bugs, each labeled with its corresponding class. Furthermore, we scanned 388 mixed volumes, each containing between 4 and 45 bugs alongside other materials. For evaluation purposes, each bug in the mixtures is manually annotated with its center position and class label. In total, 9542 volumes make up the \bugnist dataset. The numbers in each class are shown in \cref{fig:overview}. Since the bugs in each class are the same age, we expect to cover the biological variation sufficiently to be representative of the class. 

Solutions to the type of context shift, that we present here, namely a domain shift where only the context changes, and not the appearance of the objects, have wide applications in volumetric imaging. The field of volumetric imaging is dominated by medical applications \cite{oecd2023}, but there is a growing use beyond the medical area based on \eg micro-CT \cite{withers2021x,walsh2021imaging}. In all these application areas, there is a need to employ supervised deep learning to solve problems that require quantification of depicted structures. When the use case is novel and there is no directly applicable training data available --- a typical scenario in volumetric imaging of \eg materials microstructure, but also in specialized biomedical applications --- it is necessary to \emph{create} training data. If labor-intensive manual labeling can be limited or avoided, there is potentially a huge gain in productivity. Examples of applications are detecting and classifying objects in scans of marine sediments such as microplastics \cite{trusler2023x,yan2021core}, museum specimens \cite{hipsley2020high}, or detecting threat items in baggage security screening \cite{velayudhan22}.

In volumetric scans, manual labeling is significantly more demanding than in 2D. For instance, annotating a bounding box of a single object in 3D requires traversing slices of the volume and making a mental model of the object. If the volume contains densely packed complex objects, determining their bounding boxes becomes difficult. Further, bounding boxes may overlap to such an extent that they stop being a useful way to locate objects. This is why we employ center points as the annotation format in the mixtures.

The name \bugnist is a homage to the influential MNIST dataset \cite{lecun1998gradient} which has played a pivotal role in the development and evaluation of deep learning methods. Similar to the way that MNIST has been used, \bugnist is intended to be a benchmark dataset that represents a well-defined problem, where the development of solutions will make a basis for solutions that can advance the field. Specifically for \bugnist, solutions will enable object detection and classification with significantly less effort put into creating the data foundation for training. This calls for methods that can learn to predict objects in a complex context from volumes of individual objects, namely solving the \emph{context shift} problem. To our knowledge, no existing datasets have this property.  

The term \emph{context shift} refers to a shift in the data distribution between training and inference. Typical shifts in data distributions arise from changes in data acquisition systems, \eg the use of different scanners, or from variations in the objects of interest between datasets. Such problems have been addressed using domain adaptation \cite{guan2021domain,wang2018deep}. In our setting, there is almost no change in the appearance of the bugs between volumes of individual bugs and mixtures, but the change is in what surrounds the bugs, \ie the change is in the context.

We hypothesize that methods capable of learning under \emph{context shift} might also prove capable of learning from very few examples, similar to how humans learn \cite{xu2007word}. Instead of relying on seeing objects in every imaginable context, such methods could instead isolate objects from their contexts. If this hypothesis holds true, learning under context shift may reduce the need for extensive training data and the associated training costs. This is in contrast to the current trend, where larger models and datasets as seen in LLMs \cite{zhao2023survey} and computer vision \cite{kirillov2023segment,ma2024segment}, lead to better performance, but also limit who can effectively train such models. 

\paragraph{Contributions} We contribute \bugnist, a new volumetric image dataset for benchmarking domain shift. In this paper, we use \bugnist for benchmarking detection methods under context shift, trained only on individual bugs and tested on mixtures. We provide several baselines and investigate different training strategies for existing volumetric detection methods.

Our results show that volumetric object detection under context shift is difficult. We obtain reasonable results in localizing the bugs, but when a class label is assigned, there is a significant drop in performance. This is even though the classification of individual bugs is easy. Our experiments demonstrate great potential for developing new approaches for 3D object detection of complex objects.

\section{Related work}
In computer vision, the term 3D data refers to different data types, including point clouds, surfaces, volumes, and depth images. Here, we are only concerned with 3D volumes, \ie image data represented on a voxel grid.

\bugnist is created to advance deep learning methods for object detection and classification under domain shift. Especially in medical imaging, analysis of 3D volumes is in great demand. A review of medical challenges illustrates this \cite{maier2018rankings}. Their study shows that detection makes up 7\% of the tasks, classification makes up 10\% of the tasks, whereas segmentation accounts for 70\% of the tasks. Detection is important for example for diagnosis \cite{chaudhary2021detecting,harmon2020artificial,ker2019image,serte2021deep,zhou2021review}. Outside the medical domain, object detection problems include baggage security screening \cite{flitton2015object,wang2020multi,wang2021}, although the problem and the available data are still primarily in the 2D domain \cite{velayudhan22}, food inspection \cite{du2019x} such as fruit quality \cite{jarolmasjed2016postharvest}, grain classification \cite{charytanowicz2018evaluation}, foreign object detection \cite{einarsdottir2016novelty}, biological applications such as analysis of weeds \cite{wu2021deep} and other uses driven by developments in X-ray micro-CT imaging \cite{rawson2020x}, \etc 

The domain shift problem presented in \bugnist, where only the context changes, and not the appearance of the objects of interest, is not widely studied. Most research on domain shifts focuses primarily on problems where the appearance of objects changes. Solutions focus on domain adaptation, where labeled data in the source domain are available together with data without labels from the target domain \cite{pan2010domain}, or domain generalization, where the inference domain is not known at training time \cite{muandet2013domain,wang2024inter}. We focus on domain adaptation, which is also the typical use case for volumetric imaging.

Several widely used datasets are available for studying and developing methods for domain shift. For 2D image classification problems the datasets PACS \cite{li2017deeper}, Digits-DG \cite{zhou2020deep}, Office-Home \cite{venkateswara2017deep}, and DomainNet \cite{peng2019moment} developed to classify images from domains such as photos, art paintings, sketches, \etc Domain shift between artificially generated and real images \eg for segmentation is another area that receives much attention \cite{sankaranarayanan2018learning} with standard datasets such as SYNTHEIA \cite{ros2016synthia}, CITYSCAPES \cite{cordts2016cityscapes}, and GTA-5 \cite{richter2016playing}, or completely synthetic domain shift like the SHIFT dataset \cite{sun2022shift} for autonomous driving. 

In 3D, domain shift is often studied in medical applications where the same type of object is imaged with different scanning modalities \cite{li2023medical}. It can be MRI (magnetic resonance imaging), CT, OCT (optical coherence tomography), \etc Standard datasets include the CrossMoDA 2021 dataset \cite{dorent2023crossmoda} and the BRATS dataset \cite{baid2021rsna}, both of different MRI modalities. There are many other datasets used for domain adaptation for volumetric medical imaging \cite{li2023medical}, but all for adapting between modalities or between the same modality but different scanners \cite{guan2021domain}.

The structure of the imaged objects in \bugnist is different from most existing volumetric deep learning benchmark datasets. The morphological complexity of bugs is, however, representative of volumetric imaging data of \eg 3D microstructure, where there is much less use of deep learning than in \eg the medical domain \cite{withers2021x}.

Platforms like Zenodo \cite{zenodo2013cern}, TomoBank \cite{de2018tomobank}, or the Human Organ Atlas \cite{walsh2021imaging} also contain a large amount of volumetric image datasets. Different from \eg medical challenges, this data generally comes with no clear problem formulation, well-defined tasks or even labels, that allow for a comparison of methods. Further, their large volume sizes are difficult to access. It can also be difficult to judge the complexity of the data and if deep learning is needed. Some images can easily be analyzed using a threshold whereas others are complex and can greatly benefit from deep learning. Therefore, most of this data is not attractive to use as a benchmark for developing new deep learning methods. Some datasets are emerging, \eg the vessel segmentation challenge \cite{jain2023sennet}, but they remain exceptions.

There is extensive research in deep learning-based object detection, but primarily in 2D. Most methods either use a two-stage detector such as R-CNN \cite{girshick2015fast,girshick2014rich,ren2015faster}, with a region proposal and a classifier to choose among the proposed regions, or a one-stage detector such as YOLO \cite{redmon2016you}, where regions and classes are proposed in a single network. 

Object detection in volumetric images is less common. One approach has been to apply a 3D U-Net \cite{cciccek20163d} segmentation network and use each connected component of the foreground label class as a detection \cite{jaeger2020retina}. An example of using this approach is lung nodule detection in \cite{xiao2020segmentation}. Employing segmentation for object detection comes with the problem of separating objects that may be connected or combining separated segments into one object. To avoid this problem, Jaeger et al. \cite{jaeger2020retina} propose the Retina U-Net: a one-shot 3D detection algorithm that combines the Retina Net \cite{lin2017focal} and U-Net \cite{ronneberger2015u} giving detections with 3D bounding boxes. Despite the good performance of Retina U-Net, changing to a new 3D detection problem requires a time-consuming method configuration. This has been addressed in nnDetection \cite{baumgartner2021nndetection}. Inspired by the nnU-Net (no new U-Net) \cite{isensee2021nnu}, they propose a no new 3D detection method. 
On the \bugnist data, we test detection with 3D U-Net \cite{cciccek20163d}, the nnDetection method \cite{baumgartner2021nndetection}, and a 3D adaption of the Faster R-CNN model \cite{ren2015faster}.

Benchmark datasets for volumetric object detection are primarily medical. Examples include Baumgartner 
et al. \cite{baumgartner2021nndetection}, which uses 13 datasets to develop the nnDetection method. Two of these datasets were originally targeted 3D detection and include the LUNA16 for nodule detection \cite{setio2017validation}, which is a subset of the LIDC dataset \cite{armato2011lung} and contains 888 CT scans of lungs, and the ADAM dataset \cite{timmins2021comparing} for detecting aneurysms in MRI scans. ADAM is made up of 254 MRI scans. The rest of the detection datasets are medical segmentation datasets such as data from the Medical Segmentation Decathlon \cite{amber2019decathlon}, the rib fracture detection \cite{jin2020deep}, the PROSTATEx dataset \cite{armato2018prostatex}, and others. The segmentation datasets have been transformed into detection data by identifying center positions and bounding boxes for the segments. Within baggage security screening, some 3D datasets are obtainable, for example, the ALERT datasets used in \cite{wang2021}. Due to the sensitive nature of the problem, these datasets have copyright restrictions and limited availability under a non-disclosure agreement.

\section{Data acquisition and processing}

\cref{fig:overview} gives an overview of the \bugnist dataset. Each individual bug volume contains exactly one bug located near the center of the volume. The total of 9154 individual bug volumes have between 713 and 946 volumes of each class distributed across 12 bug classes (species and stages). Bug mixtures contain several bugs, leaves, cotton, shredded paper, and wood shaving. There are in total 388 mixtures. The mixtures contain 4 to 45 bugs, totaling 5011 bugs with an average of 13 bugs per mixture, see \cref{fig:mixes_summary}. In bug mixtures, we have manually marked the center positions of each bug and provided it with a class label. 

All scans were acquired using a laboratory micro-CT scanner (Nikon metrology scanner) that was set to record volumes of 2000 voxels cubed with an isotropic voxel size of 26.62~µm. Being able to acquire such large volumes allowed us to scan multiple bugs and bug mixtures at once, and then crop them into smaller volumes. The dataset was collected through nearly 250 scans, each taking 30 minutes, over 90 days. Care has been taken that the scanning parameters stay unchanged for all scans, and we have calibrated the scanner daily.

All bugs were placed in tubes, either as individuals or in mixtures. We used cotton, which is almost transparent to X-rays, to separate the bugs, either as individual bugs or mixtures. For the mixtures, we varied the difficulty of the bug detection task by adding leaves, shredded paper, or wood shavings to some of the mixtures, and also varied the number of bugs, their class, and the packing density.

The cylindrical shape of the tubes was easy to segment using a circular Hough-transform \cite{xie2002new} and sparse layered graphs \cite{jeppesen2020sparse}, leaving us with a cylinder of only bugs and cotton. Due to the low intensity of cotton compared to the bugs, we could easily divide each cylinder into volumes containing one bug each. The volumes have been manually inspected and cleaned to correct cases where something went wrong, \eg parts of another bug showing in the periphery of the volume. The final volumes of scans of individual bugs are $450 \times 450 \times 900$ voxels, the size which was chosen to accommodate the biggest bugs. The mixtures were segmented similarly and the resulting volumes are $650 \times 650 \times 900$ voxels. All volumes are stored as 8-bit unsigned integers with their original spatial resolution of 26.62~µm (physical voxel size). Our dataset comprises 1.8 trillion voxels, approximately a quarter of the RGB values in ImageNet.

\input{images/mixes_summary}  

The bugs scanned for the \bugnist dataset are bred as pet food and have been frozen before being packed for scanning. Despite careful packing, some scanned bugs may appear damaged, \eg a few crickets have broken off antennas or miss a leg. We have chosen to accept these defects because they reflect some variation in a population and add to the challenge of the dataset. 

To create additional lightweight datasets, we created four downsized versions of all volumes. The individual bug volumes, originally 900 voxels long, have lengths of 512, 256, 128, and 64 voxels in the downscaled versions, and all the other lengths are scaled accordingly. The voxel size of the coarsest data is 374.3~µm, which is a scaling factor of approximately 14 compared to the original scan.

\section{Experiments}

To establish baselines for volumetric detection under context shift (domain shift with no change in object appearance), we have selected and adapted three existing detection models. Furthermore, we have applied three training strategies, yielding a total of nine baseline models. The diversity in model choices and training strategies contributes to a comprehensive set of baseline models that provide a strong foundation for object detection under context shift. We briefly describe the selected methods and how they were trained. Full details are provided in the supplementary material. To determine the difficulty of differentiating between the individual bugs, we also performed a classification experiment which is described in the supplementary material.

All experiments are carried out using the x128 data, which is a compromise between 
memory needs and the preservation of structural details of the bugs.

\subsection{Training strategies for detection under context shift}
\begin{figure}
\begin{center}
\input{images/illustration3}
\end{center}
\caption{The top line shows the automated annotation pipeline for individual bugs. By thresholding, bounding boxes and segmentation masks are easily obtained. The bottom shows the synthetic mixtures used for the more advanced training strategy. }
\label{fig:process}
\end{figure}

To provide a simple baseline investigation for the \bugnist dataset, we investigate some strategies for solving the detection problem under context shift. Since the purpose of our paper is to initiate research in domain adaptation methods for volumetric data under context shift, this baseline study is intended as an offset for future developments based on the \bugnist dataset. Our first attempt is to train models from the volumes of individual bugs. Our two other attempts are to use a data augmentation strategy by creating synthetic mixtures. Data augmentation has previously been shown to greatly benefit domain shift problems \cite{wang2024inter}. The automated annotation of the \bugnist data and the data that we use for training detection algorithms are illustrated in \cref{fig:process}.

\paragraph{Single bugs} As a first stage, we train the models to detect the bugs in the individual scans. For models requiring ground truth segmentation labels, those are generated by thresholding the volumes and only keeping the largest connected component. The value of the class label then replaces the foreground voxels. This creates segmentation labels where each voxel has one of 13 values including background or one of the twelve bugs. For models requiring ground truth bounding boxes, those are extracted from each connected component. This process of preparing the training data using this simplest strategy is illustrated in the top line of \cref{fig:process}.

\paragraph{Synthetic mixtures} As a second stage, we train the models on artificial mixes. These are generated by randomly sampling individual bug scans and adding them to a common volume. We start with an empty volume of $256^3$ and we set a random target number of bugs drawn from a normal distribution, $\mathcal{N}(32, 6)$. A random bug from a random class (both uniform distributions) is drawn. To create variation, the selected sample is flipped around a random axis, rotated a random multiple (0-3) of 90 degrees around the first axis, and then transposed randomly. The bug volume is intensity normalized, smoothed with a Gaussian filter with $\sigma = 0.5$, and thresholded at $0.2$. We draw a random voxel position from the volume and only insert the bug if its bounding box does not exceed the volume boundaries nor overlaps with existing labeled voxels. We attempt a maximum of 1000 random positions to achieve a valid insertion before the bug is discarded. In total 300 synthetic mixtures are created.

\paragraph{Crowded synthetic mixtures} Finally, since the bugs are packed very close together in the real scans, we create crowded synthetic mixtures. Here the bugs are packed closer by reducing the volume size to $200 \times 200 \times 200$, adjusting the target number of bugs to $\mathcal{N}(64,8)$, and relaxing the insertion criteria such that no labeled voxels can be overwritten. This effectively allows for a larger overlap between bounding boxes and that bugs can touch each other. Again, a total of 300 synthetic mixtures are created.

\subsection{Baseline detection models}
Despite important uses of detecting objects in 3D volumes, the number of methods addressing volumetric object detection is much smaller compared to 2D, and only a few methods are readily available for volumes. So, to establish baselines for detection under context shift, we have selected and adapted three detection approaches. They vary in how they utilize training data; relying on either ground truth segmentation labels, ground truth bounding boxes, or both. 

\paragraph{Detection U-Net} Inspired by \cite{baumgartner2021nndetection} we use U-Net \cite{cciccek20163d} as one baseline detection model. U-Net is a dominating model for volumetric segmentation, making it a natural choice for adapting into a baseline detection model. Detection U-Net is trained on volumes accompanied by ground truth segmentation labels. It is first trained to predict the segmentation of the single bugs. Then, starting from the obtained weights, it is trained to predict the segmentation of the synthetic mixtures. Finally, again starting from the previous weights, it is trained to predict the segmentation in the crowded synthetic mixtures. For all training runs, we augment the images with random rotations, translations, geometric scaling, and intensity scaling. We train the models using the sum of Dice and cross entropy loss. During training we hold out a validation split which we evaluate on after every epoch. At the end of training, the model which scored the highest on the validation split is kept. Like \cite{baumgartner2021nndetection}, we extract bounding boxes by thresholding the softmax probabilities (at 0.75) and then extracting the connected components for each channel. Only components larger than 125 ($5^3$) voxels are kept. We use the U-Net implementation from MONAI \cite{cardoso2022monai}.

\paragraph{Faster R-CNN} As an example of region proposal networks, we have chosen Faster R-CNN \cite{ren2015faster} with the modifications from \cite{li2021benchmarking}. We have adapted the implementation from torchvision \cite{torchvision2016} to allow for 3D detection.  We train in the same sequence as the U-Net: single bugs, synthetic mixtures, and crowded synthetic mixtures.  We use the same augmentations as for the U-Net. In addition, we also randomly perturb the bounding box corners. The model is only trained on bounding boxes and does not make use of the voxel labels. The model is trained with smooth L1 loss for bounding box regression and cross-entropy for objectness and class predictions. For full details, we refer to \cite{ren2015faster, li2021benchmarking}. As with U-Net, we hold out a validation split which is evaluated after each epoch. After training, the model scoring the highest on the validation split is kept. For inference, only bounding boxes with an objectness score over 0.05 are kept.

\paragraph{nnDetection} The third detection model, nnDetection, trains the RetinaUNet \cite{jaeger2020retina} detection model using a combination of segmentation masks and bounding boxes. The bounding box branch is trained using a generalized IoU loss for regression and cross-entropy for classification. The voxel segmentation branch is trained with a sum of Dice and cross-entropy loss. As nnDetection is a self-configuring framework, we refer to \cite{baumgartner2021nndetection,isensee2021nnu} for further training and model details. We simply provide the data as instructed. We train a separate model for single bugs, synthetic mixtures, and crowded synthetic mixtures. nnDetection trains using 5-fold cross-validation, and the final predictions are made using an ensemble over the best model from each fold \cite{baumgartner2021nndetection}.

\subsection{Evaluation}
We assess model performance by evaluating their detection accuracy on real scans of densely packed bugs, using manually annotated center points. These annotations are matched with predicted bounding boxe centers based on Euclidean distance, optimized using Munkres' algorithm \cite{munkres1957algorithms}. Matches are only accepted if the annotated point falls within a $1.5 \times$ inflated bounding box, as illustrated in \cref{fig:match-illustration}. This inflation ensures a fair assessment for detections like in \cref{fig:match-illustration}(b). This, combined with matching on the bounding box centers, helps ensure that large boxes are not favored since it places more weight on the location of the boxes. We compute precision to avoid favoring overdetections, recall to measure the proportion of correct matches, and F1-score to combine the two. Further, we first compute the metrics purely from the point matches without considering classes --- thereby evaluating if the models can detect the presence of bugs vs. background material. Next, we recompute the matches where we only allow matches between points of equal class.

\input{images/match_illustration}

\section{Results}

The results are summarized in \cref{tab:det_results}, example results are shown in \cref{fig:mix_detection}, and the distributions of the metrics without class information are shown in \cref{fig:metric_hists}. U-Net performs relatively well, independent of what it was trained on, whereas Faster R-CNN performs worse on crowded mixtures and nnDetection on single bugs. If class labels are required, the performance is generally low.

The distributions in \cref{fig:metric_hists} show that U-Net has a narrower spread around the mean for the F1-score, whereas the two others are spread more out. Wrt. the precision, many detections obtained by the nnDetection method are 1, but have a lower recall, meaning that those that it predicts are correct, but there are many potential detections that it misses. It is the opposite for the Faster R-CNN, where many datasets get a recall of 1, but a lower precision, meaning that it detects many of the bugs, but also has many false detections.

To investigate the capabilities of the three methods without the domain shift between the synthetic and real mixtures, we included the results in \cref{tab:train_det_results}. This table shows the results of the validation data for training the models on the volumes of individual bugs and the two synthetic mixtures with the class labels. Since we have the bounding boxes for these datasets, we compute the mean average precision (mAP), similar to, \eg, COCO detection  \cite{lin2014microsoft}. We compute mAP with the IoU threshold running from 0.5 to 0.95 (shown as mAP) and fixed at 0.5 (mAP@50). We note that the values are not directly comparable with those in \cref{tab:det_results}. We show example detections in the supplementary. The U-Net generally has high performance, Faster R-CNN has high performance for single bugs but has difficulty with the mixtures, whereas nnDetection performs better in the mixtures than for the individual bugs. Generally, the detection performance of bugs is far from perfect and can be improved, also when training and testing within the same domain.

\input{tables/bugnist_detection_results}
\input{images/mix_detection}
\input{images/metric_hists}
\input{tables/train_task_results}

\section{Discussion}

The domain shift introduced by shifting the context surrounding the bugs significantly increases the difficulty of the analysis problem. While U-Net and nnDetection achieve high performance on their training tasks (at least for synthetic mixtures), they struggle when transferred to real densely packed scans. It is also interesting that the domain shift has such an effect on the ability of the models to infer the correct class even when accurately localizing the bug. This is surprising since detecting the bugs in the training domain can be done with high accuracy, as shown in
\cref{tab:train_det_results}, and pure classification (see supplementary) gave an accuracy of 0.97. Despite the models for classification being different than the detection models, these results show that even when the objects do not change appearance, the shift in the domain by changing the context surrounding the objects is sufficient for deep learning-based detection models to lose performance.

In preparing the data, we have strived to get the scans of individual bugs and bugs in mixtures to have the same appearance. Since bugs are flexible, there will be differences in the pose of the bugs in mixtures and individual bugs. Larvae will sometimes bend differently and bugs with antennae or legs will bend differently. Also, the cotton is visible and can slightly influence the appearance. Despite these differences, the voxel intensity distributions in individual scanned bugs and mixtures are the same, and therefore the \bugnist dataset is a good basis for developing detection methods that can handle context shift. If the mixtures were (partly) labeled for segmentation, \bugnist could also be used to investigate segmentation algorithms under context shift. 

We chose to annotate only the center positions of the bugs. As the bugs are often elongated and at odd angles, axis-aligned bounding boxes would contain significant empty space and overlap between neighboring boxes---thus offering limited benefit over points. This could be solved with \emph{object} aligned bounding boxes or full voxel annotations. However, this would drastically increase the annotation time. We chose to prioritize having a large number of annotations such that the test set can have a large variety of configurations. Furthermore, given that current methods show ample room for improvement (\cf \cref{tab:det_results}), we find that point annotations combined with the chosen metric are sufficient to measure method performance and improvements of future approaches.

The methods and experiments are chosen as a basic baseline to illustrate the properties of the \bugnist dataset. From the experiments, we learned that data augmentation by composing synthetic mixtures is a feasible approach for improving detection performance. More investigations into packing algorithms and ways to combine the synthetic mixed bugs with other backgrounds can potentially improve performance further. But it is also clear that there is great potential for developing new methods for object detection in volumes. We chose Detection U-Net, nnDetection, and our adaption of Faster R-CNN to 3D volumes because these methods were available, widely used, and are based on complementary ideas. We did not find more recent alternatives for volumetric analysis.

Our results show that Faster R-CNN performs less than the two other methods. While it scores highest in \cref{tab:det_results}, it is clear from \cref{fig:mix_detection} that the bounding boxes are of much lower quality. Furthermore, it does not score well on the training tasks, except when trained on individual bugs. To ensure the poor performance of Faster R-CNN was not just the result of too few training volumes, we redid the experiments with 1000 synthetic mixes instead of the original 300. Performance remained the same.  It is also noteworthy that nnDetection scores overall lower than U-Net on both the training and evaluation tasks, even though it also has access to the segmentation masks for training.

Finally, \bugnist can have other valuable use cases than what we have explored in this paper. It could be used for exploring segmentation methods, generative models, or image registration. The good contrast makes it possible to extract surface meshes for use in geometric deep learning. The fact that we have scanned many specimens of each species also allows for investigating methods for analyzing the morphological variation of the scanned bugs. 

\section{Conclusion}
With our \bugnist benchmark dataset for domain shift in volumetric data, we demonstrate how detecting objects, that have the same appearance in the source and target domain, fails when the objects in the two domains are placed in different contexts. By contexts, we refer to the appearance of the image surrounding the objects. To our knowledge, the \bugnist dataset is the only dataset designed for developing methods for context shift, \ie domain shift where the objects stay the same but the surroundings change. Solutions will enable an automated annotation of data for volumetric object detection, which is an area that has not received much attention. There are, however, a range of applications where object detection is needed, including medical imaging and emerging areas such as micro-CT scanning of materials. The \bugnist dataset can also be used for investigating other problems such as segmentation under domain shift provided reference annotations are created.

\section*{Acknowledgments}
We acknowledge the 3D Imaging Center at DTU, the Infrastructure for Quantitative AI-based Tomography (QUAITOM), supported by the Novo Nordisk Foundation (grant number NNF21OC0069766) and STUDIOS: Segmenting Tomograms Using Different Interpretation of Simplicity funded by the Villum Foundation (grant number VIL50425).

%
%

\title{Supplementary material for:\\
\bugnist -- a Large Volumetric Dataset for\\ Object Detection under Domain Shift}

\author{Patrick Møller Jensen\orcidlink{0000-0002-8479-4885} \and
Vedrana Andersen Dahl\orcidlink{0000-0001-6734-5570} \and
Rebecca Engberg\orcidlink{0009-0005-9546-1954} \and
Carsten Gundlach\orcidlink{0000-0002-2895-1882} Hans Marin Kjer\orcidlink{0000-0001-7900-5733} \and
Anders Bjorholm Dahl\orcidlink{0000-0002-0068-8170}}
\authorrunning{P.M.~Jensen et al.}

\institute{Technical University of Denmark, Kgs. Lyngby, Denmark\\
\email{\{patmjen,vand,reen\}@dtu.dk,cagu@fysik.dtu.dk,\{hmjk,abda\}@dtu.dk}}

\maketitle

\section{Introduction}
In this supplementary material, we show how data was collected, present examples from the dataset for a better understanding of the appearance of the volumes, provide implementation details for the baseline methods, and show the results of classification experiments and example detections. Upon publication, the material covered in this document will be made available on our webpage together with the data and the code for the analysis.

\begin{figure}[!b]
    \centering
        \includegraphics[height=44mm]{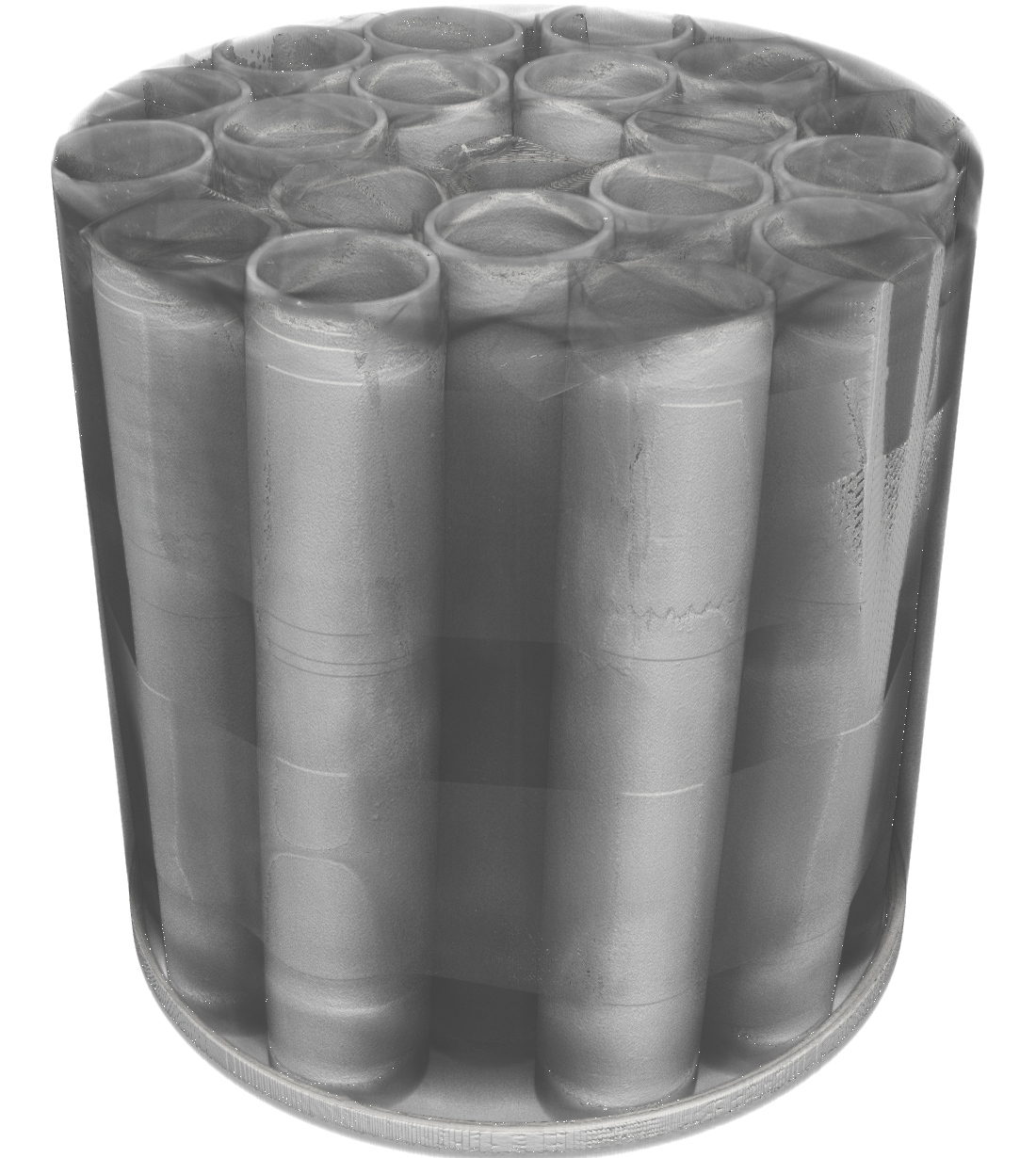}
        \includegraphics[height=44mm]{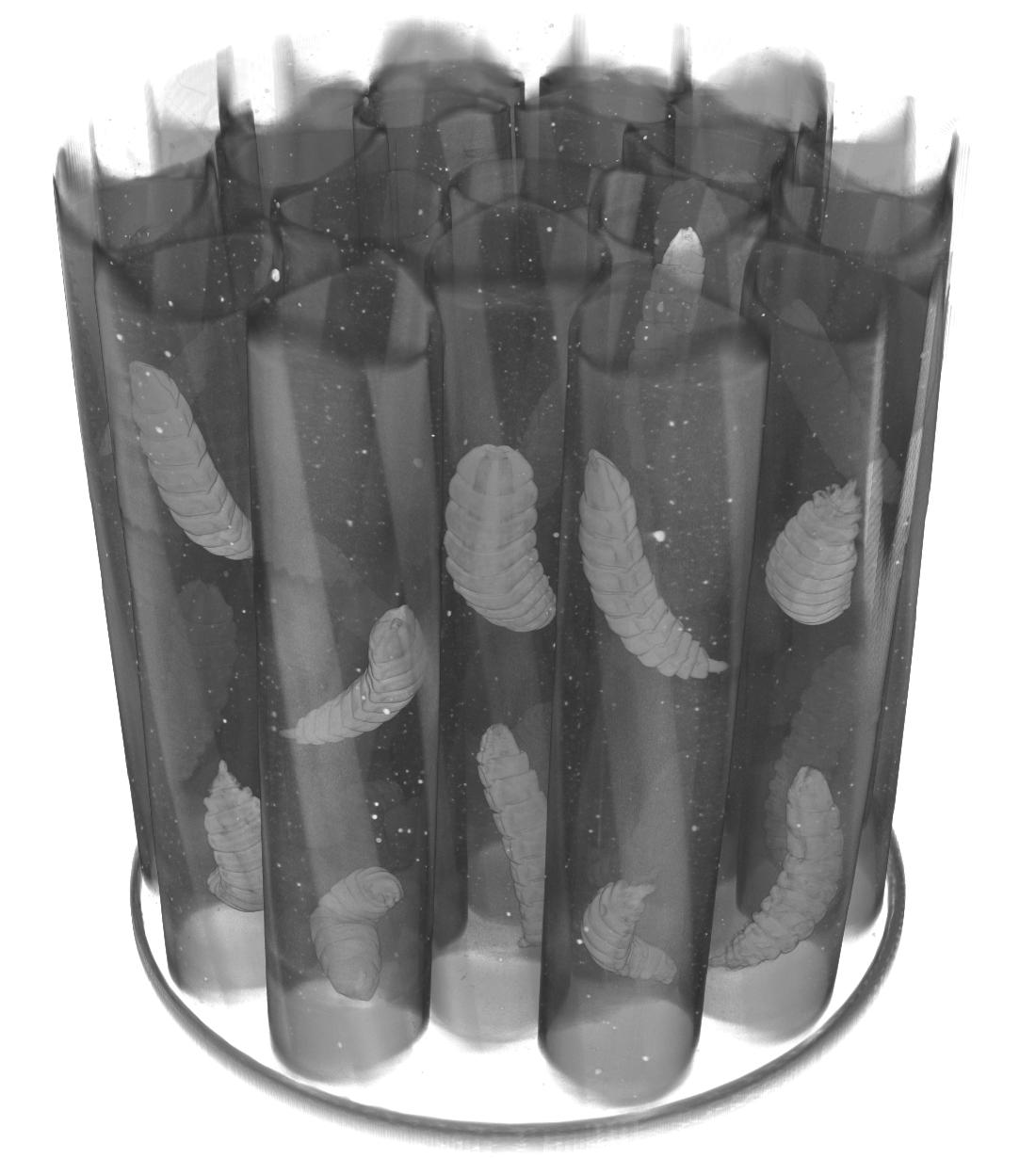}
        \includegraphics[height=44mm]{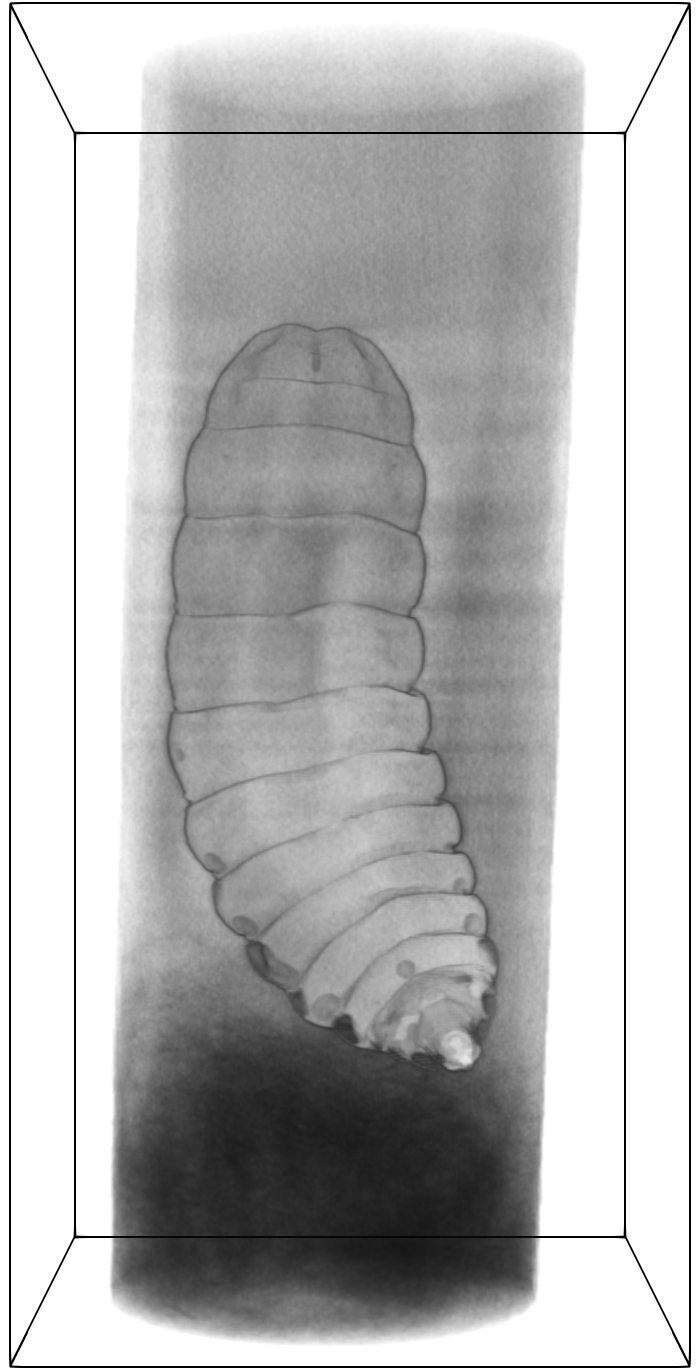}
    \caption{Volume renderings of the full scans and a single bug. The left shows the scanned bundle of tubes and the middle shows the same volume with higher transparency to reveal the bugs inside the tubes. The right shows a rendering of a single bug with low transparency to make even the air around the bug slightly opaque and highlights the crop-out cylindrical region containing a bug, air, and some cotton at the bottom. 
    The region outside the crop-out cylinder is set to zero and is fully transparent.\label{fig:setup}\vspace{-30pt}}
\end{figure}

\section{Data acquisition}
Volume renderings of the full scans are shown in \cref{fig:setup}. Bugs are placed in tubes, and inside the tubes, the bugs are separated by cotton. Cotton is chosen because it does not attenuate X-ray much and gives low voxel intensities in the volumes. Therefore, the cotton does not affect the image much. We have chosen to place the bugs in tubes since this allows us to scan many at once, and we can easily identify the tubes and the bugs.

\section{Data overview}
To make the \bugnist available for different uses, we provide the dataset in different resolutions. The volume sizes and the total file sizes are listed in \cref{tab:image_sizes}. Illustrations of the downscaled images are shown in \cref{fig:bug_examples_down_scaled}. We have also included illustrations of the scans of mixed bugs in \cref{fig:mixture_examples_down_scaled}.

\begin{table*}[!t]
    \centering
    \small
    \caption{Volume sizes and file sizes (zipped files) of the \bugnist datasets for different downscaled versions.}
    \setlength{\tabcolsep}{15pt}
    \begin{tabularx}{\linewidth}{Xrrr}
    \toprule
    & \multicolumn{2}{c}{Volume sizes} & \\
    Scale & Individual volume & Mix volume & \\
    \midrule
    x64 & $64\times32\times32$ & $64\times 46\times 46$ & \\
    x128 & $128\times32\times32$ & $128\times92\times92$ & \\
    x256 & $256\times128\times128$ & $256\times185\times185$ & \\
    x512 & $512\times256\times256$ & $512\times370\times370$ & \\
    x900 & $900\times450\times450$ & $900\times650\times650$ & \\
    \midrule
    & \multicolumn{3}{c}{Dataset file sizes} \\
    Scale & Individual bugs &  Mixtures & Total \\
    \midrule
    x64 & 104 MB & 33.7 MB & 138 MB \\
    x128 & 684 MB & 163 MB & 847 MB \\
    x256 & 5.5 GB & 1.8 GB & 7.3 GB \\
    x512 & 45.0 GB & 14.6 GB & 59.6 GB \\
    x900 & 245 GB & 47.5 GB & 292.5 GB \\
    \bottomrule
    \end{tabularx}
    \label{tab:image_sizes}
\end{table*}

\begin{figure*}
    \centering
    \input{images/illustration3_512_64.tex}
    \captionof{figure}{Overview of the downscaled \bugnist individual bugs with sizes  x512, x256, x128, and x64. Images are maximum projections from the side (first row of each block) and from the top (second row of each block). These images were used for classification experiments also included in this supplementary material.}
    \label{fig:bug_examples_down_scaled}
\end{figure*}

\begin{figure*}
    \centering
    \input{images/illustration4.tex}
    \captionof{figure}{Examples of the downscaled \bugnist mixtures containing multiple bugs and other material. Images are maximum projections from the side (first row of each block) and from the top (second row of each block). We have kept track of the number and type of bugs in all mixed volumes, and a subset of 45 mixed volumes has been annotated for evaluation of the detection methods. }
    \label{fig:mixture_examples_down_scaled}
\end{figure*}

\section{Details for detection baselines}
We here provide details about training the baseline detection models.
\paragraph{U-Net} For all experiments, we use the U-Net implementation from MONAI \cite{cardoso2022monai}. The U-Net has four levels. Each block in the encoder and decoder consists of two residual units. The convolutions use 32 channels for the first level, 64 for the second, 128 for the third, and 256 for the fourth. All remaining parameters are at their default values as of MONAI version 1.1.0. 

For the individual bugs, we train on the 128x version where the volumes are zero-padded to be $128^3$ and normalized to have intensities in $[0, 1]$. To extract a mask, we smooth using a Gaussian kernel with $\sigma = 2$ voxels, threshold at 0.25, and keep the largest connected component. We also perform random rotations (angle 0-$\pi$ radians), axis flips, translations ($\pm$ 128 voxels), spatial scaling (0.9x-1.1x), and intensity scaling (0.75x-1.25x) as data augmentation. The network is trained using the Adam \cite{kingma2014adam} optimizer with a learning rate of $10^{-4}$, $\beta_1 = 0.9, \beta_2 = 0.999$ and a batch size of 4. The loss is the sum of Dice and cross-entropy loss, as recommended in \cite{isensee2021nnu}. The network is trained for 48 hours  (roughly 800 epochs). We hold out 40\% of the training volumes for validation.

For the synthetic mixtures, we train on the generated volumes and pre-computed label masks. We use the same data augmentation as for the individual bugs, except that we skip random translations and only rotate $\pm \pi/8$ radians. The batch size is also decreased to 2. All remaining details are the same as for the individual bugs. Again, the networks trained for 48 hours (roughly 800 epochs for the synthetic mixes and 5000 epochs for the crowded synthetic mixes). Finally, for the synthetic mixes, we initialize the weights with the values achieved after training on the individual bugs. For the crowded synthetic mixes, we initialize the weights with those achieved from training on the non-crowded synthetic mixes.

\paragraph{Faster R-CNN} For all experiments, we use the faster R-CNN implementation from torchvision \cite{torchvision2016} --- specifically, \texttt{fasterrcnn\_resnet\_fpn\_v2} --- which we manually adapt to 3D volumes. All parameters are at their default values as of torchvision version 1.14.1+cu116.

For the individual bugs, all training details are the same as for the U-Net. The loss, now only based on bounding boxes, is a smooth L1 loss for the bounding box regressions and cross-entropy for objectness and class predictions. See \cite{li2021benchmarking,ren2015faster} for full details. We extract bounding boxes for training from the mask volumes computed with the same strategy as for the U-Net. Note that the bounding box is extracted \emph{after} augmenting the image, which provides additional augmentation. The model is trained for 24 hours, totaling roughly 200 epochs, which was sufficient to learn the training task. For inference, only bounding boxes with an objectness score over 0.05 are kept.

\begin{figure}
    \centering
    \begin{subfigure}{0.45\linewidth}
        \centering
        \includegraphics[width=\linewidth]{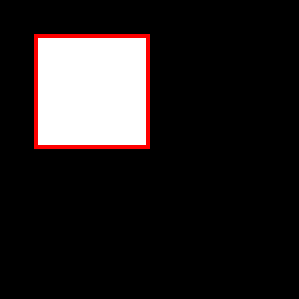}
        \caption{Before augmentation.}
    \end{subfigure}
    \begin{subfigure}{0.45\linewidth}
        \centering
        \includegraphics[width=\linewidth]{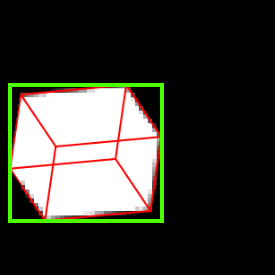}
        \caption{After augmentation.}
    \end{subfigure}
    \caption{Re-computing bounding boxes after augmentation. The original bounding box is shown in red (a). After augmentation (b), the recomputed bounding is shown in green. It is given by the axis-aligned bounding box of the augmented original box.}
    \label{fig:bbox-aug}
\end{figure}

For the synthetic mixes, training details are again the same as for the U-Net, including the weight initialization strategy. We train on the generated volumes and pre-computed bounding boxes (boxes are computed at volume generation time). To ensure bounding boxes remain axis-aligned, we re-compute the boxes after augmentation as illustrated in \cref{fig:bbox-aug}. Additionally, we randomly perturb bounding box corners with the Gaussian noise of std. dev. 1 voxel. Both models were trained for 24 hours (roughly 1600 and 2900 epochs for the non-crowded and crowded mixes respectively). After this, additional training did not improve the performance.

\paragraph{nnDetection} We use the nnDetection implementation from the official GitHub\cite{nndetectGitHub}, specifically commit d637c5e2da16e0fe7cf8a5b860907eb57e60d4fe. All parameters are kept at their default values. For full details, we refer to \cite{baumgartner2021nndetection} and the GitHub repository, but we provide a brief description here. The detection model is supervised by both image label masks and bounding boxes. The bounding box branch is supervised by a generalized IoU loss for regression and cross-entropy for classification. The voxel segmentation branch is supervised by a sum of Dice and cross-entropy like the U-Net model. We supply the images and labels and perform training as described in the README. Different from U-Net and Faster R-CNN, we do not reuse weights. First, nnDetection analyzed the provided data to create a training plan and optionally resample the images. Then, it was trained using a 5-fold cross-validation. The folds are used to optimize hyperparameters, such as the threshold used for non-maximum suppression. After training, the cross-validation results are consolidated, and inference is performed using an ensemble.

\section{Classification of individual bugs}
The performance obtained in our detection experiments shows that detecting and classifying individual bugs from mixtures is very difficult. We get significantly better results when just detecting the location of bugs, and this might suggest that bugs are difficult to classify. This is, however, not the case. We have made an extensive classification experiment of the volumes of individual bugs, and as shown in \cref{tab:clf_results}, we get an almost perfect classification. This underlines the fact that the change in context from volumes of individual bugs to mixtures makes detection very difficult.

\begin{table*}[b]
    \centering
    \small
    \caption{Classification results on the \bugnist test data for individual bugs.}
    \setlength{\tabcolsep}{2pt}
    \def\catspace{20pt}
    \begin{tabularx}{\linewidth}{Xrr@{\hskip \catspace}rr@{\hskip \catspace}rr@{\hskip \catspace}rr}
        \toprule
        BugNIST & \multicolumn{2}{l}{x64} & \multicolumn{2}{l}{x128} & \multicolumn{2}{l}{x256} & \multicolumn{2}{l}{x512} \\
        Model & Acc. & AUC & Acc. & AUC & Acc. & AUC & Acc. & AUC \\
        \midrule
        DenseNet121 \cite{huang2017densely} & 0.94 & 0.998 & 0.96 & 0.999 & 0.98 & 0.999 & 0.97 & 0.999  \\
        DenseNet169 \cite{huang2017densely} & 0.95 & 0.998 & 0.97 & 0.999 & 0.97 &  0.999 & 0.98 & 0.999 \\
        DenseNet201 \cite{huang2017densely} & 0.95 & 0.998 & 0.96 & 0.999 & 0.98 & 0.999 &    - &    - \\
        \midrule
        ResNet18 \cite{he2016deep} & 0.94 & 0.998 & 0.96 & 0.998 & 0.97 & 0.999 &    - &    - \\
        ResNet34 \cite{he2016deep}  & 0.94 & 0.997 & 0.97 & 0.999 & 0.98 & 0.999 &    - &    - \\
        ResNet50 \cite{he2016deep}  & 0.95 & 0.998 & 0.96 & 0.999 & 0.98 & 0.999 &    - &    - \\
        \midrule
        SEResNet34 \cite{hu2018squeeze} & 0.93 & 0.997 & 0.94 & 0.997 & 0.97 & 0.998 &   - &  - \\
        SEResNet50 \cite{hu2018squeeze} & 0.92 & 0.996 & 0.96 & 0.998 & 0.97 & 0.999 &   - &  - \\
        \midrule
        ConvNext-T \cite{liu2022convnet} & 0.82 & 0.983 & 0.90 & 0.993 &    - &     - &    - &     - \\
        ConvNext-B \cite{liu2022convnet} & 0.84 & 0.988 & 0.90 & 0.994 &    - &     - &    - &     - \\
        \midrule
        ViT \cite{dosovitskiy2021image} & 0.86 & 0.979 & 0.87 & 0.980 &    - &     - &    - &     - \\
        SwinT-v2-T \cite{liu2022swinv2} & 0.88 & 0.987 & 0.86 & 0.984 &    - &     - &    - &     - \\
        SwinT-v2-S \cite{liu2022swinv2} & 0.90 & 0.992 & 0.90 & 0.993 &    - &     - &    - &     - \\
        \bottomrule
    \end{tabularx}
    \label{tab:clf_results}
\end{table*}

We have investigated six classification methods: DenseNet \cite{huang2017densely}, ResNet \cite{he2016deep}, SEResNet50 \cite{hu2018squeeze}, Vision Transformers \cite{dosovitskiy2021image} from MONAI \cite{cardoso2022monai}, Swin Transformers \cite{liu2021swin,liu2022swinv2} from torchvision \cite{torchvision2016}, and ConvNext \cite{liu2022convnet} from torchvision \cite{torchvision2016}. The torchvision models were adapted to 3D by us. These models are selected because they are well-known for their high performance and span both older tried-and-tested methods and new state-of-the-art models. We train the models on the volumes of individual bugs to predict the bug species. After every epoch, we evaluate the models on the validation set and keep the best model. After training, we evaluate the model on the test set. 

We show accuracy and AUC metrics for the classification results in \cref{tab:clf_results}. Due to computational limitations, we only trained a few models on the largest versions. However, the smaller datasets still give a good assessment of the models, as they perform consistently well over the dataset sizes.  



\begin{figure}[ht]
    \centering
    \includegraphics[width=1\linewidth]{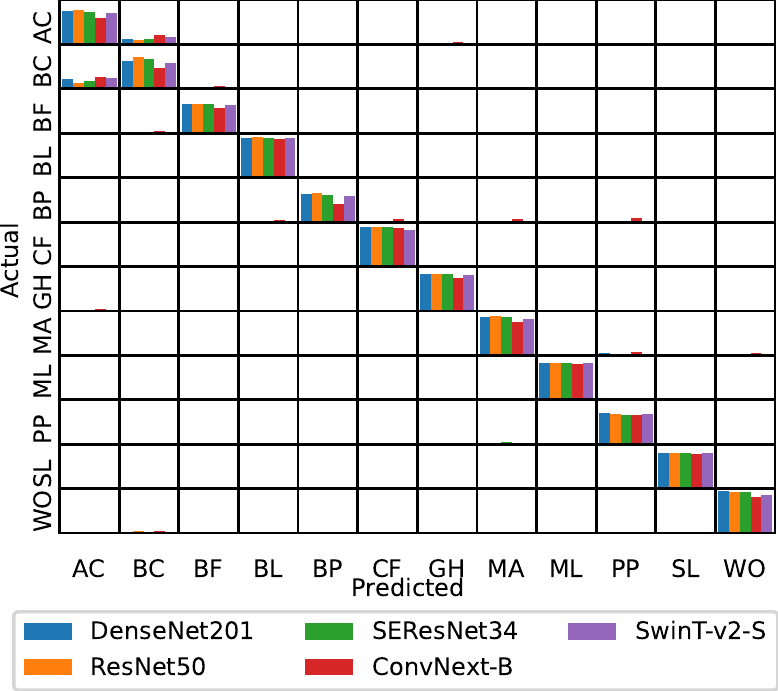}
    \caption{Confusion matrix for classifying individual bugs from BugNISTx64.}
    \label{fig:confmat}
\end{figure}
We also show a confusion matrix in \cref{fig:confmat} for a selection of models on the BugNISTx64 data. It is only the fine-grained classification of brown and black crickets (classes AC and BC) that causes issues. Looking at just these classes, the accuracy is between 0.70 and 0.86.

The classification results were left out of the main material for the \bugnist dataset, since the classification proved to be a very easy problem.



\subsection{Classification training}
As default, we train the DenseNet \cite{huang2017densely}, ResNet \cite{he2016deep}, and SEResNet \cite{hu2018squeeze} models for 200 epochs on the x64, x128, and x256 datasets and 100 epochs on the x512 datasets. The models are trained with the Adam \cite{kingma2014adam} optimizer using a learning rate of 1e-4 and other parameters as suggested by \cite{kingma2014adam}.

The ConvNext \cite{liu2022convnet}, ViT \cite{dosovitskiy2021image}, and SwinT-v2 \cite{liu2022swinv2} models were trained for 500 epochs as we found that they needed longer to converge. The models are trained with AdamW \cite{loshchilov2018decoupled} optimizer with weight decay of 0.05, a learning rate of 1e-4 for ConvNext, and 1e-6 for ViT and SwinT-v2. Other parameters are as suggested by \cite{loshchilov2018decoupled}. For the learning rate, we additionally employ a linear ramp-up over the first 60 epochs starting from 10\% of the final value followed by a cosine decay.

All models were trained using cross-entropy as the loss function. For data augmentation, we use random axis flips and random zooms (factor 0.9 to 1.1). On the x64 and x128 datasets, the batch size was set to 32, and models were trained on a single NVIDIA V100 32GB GPU. On the x256 and x512, the batch size was set to as large as possible, and models were trained on a single NVIDIA A100 80GB GPU. See \cref{tab:model_training} for specific values.
Finally, due to our computational budget, we limited the run time of the training to 24 hours. Therefore, the exact number of training epochs may differ from the default for the x256 and x512 datasets. We show specific values for all runs in  \cref{tab:model_training}.

\begin{table}[ht!]
    \centering
    \small
    \setlength{\tabcolsep}{2pt}
    \def\catspace{10pt}
    \caption{Training parameters for BugNIST classification. BS refers to batch size, and Eps. to the number of epochs trained.}
    \label{tab:model_training}
    \begin{tabularx}{\linewidth}{Xrr@{\hskip \catspace}rr@{\hskip \catspace}rr@{\hskip \catspace}rr}
        \toprule
        BugNIST & \multicolumn{2}{l}{x64} & \multicolumn{2}{l}{x128} & \multicolumn{2}{l}{x256} & \multicolumn{2}{l}{x512}  \\
        Model & BS & Eps. & BS & Eps. & BS & Eps. & BS & Eps. \\
        \midrule
         DenseNet121 \cite{huang2017densely} & 32 & 200 & 32 & 200 & 8 & 200 & 4 & 80 \\
         DenseNet169 \cite{huang2017densely} & 32 & 200 & 32 & 200 & 8 & 200 & 4 & 125 \\
         DenseNet201 \cite{huang2017densely} & 32 & 200 & 32 & 200 & 8 & 200 & - & - \\
        \midrule
         ResNet18 \cite{he2016deep} & 32 & 200 & 32 & 200 & 8 & 200 & - & - \\
         ResNet34 \cite{he2016deep} & 32 & 200 & 32 & 200 & 8 & 200 & - & - \\
         ResNet50 \cite{he2016deep} & 32 & 200 & 32 & 200 & 8 & 200 & - & - \\
        \midrule 
         SEResNet34 \cite{hu2018squeeze} & 32 & 200 & 32 & 200 & 8 & 200 & - & - \\
         SEResNet50 \cite{hu2018squeeze} & 32 & 200 & 32 & 200 & 8 & 200 & - & - \\
        \midrule
         ConvNext-T \cite{liu2022convnet} & 32 & 500 & 16 & 500 & - & - & - & - \\
         ConvNext-B \cite{liu2022convnet} & 32 & 500 & 16 & 357 & - & - & - & - \\
        \midrule 
         ViT \cite{dosovitskiy2021image} & 32 & 500 & 32 & 500 & - & - & - & - \\
         SwinT-v2-T \cite{liu2022swinv2} & 32 & 500 & 32 & 500 & - & - & - & - \\
         SwinT-v2-S \cite{liu2022swinv2} & 32 & 500 & 16 & 379 & - & - & - & - \\
         
        \bottomrule
    \end{tabularx}
\end{table}

\section{Example detections on training volumes}

To qualitatively assess the performance of the detection baselines on the task they were trained on, we show selected examples of the detection in \cref{fig:mix_detection}. 
The figure illustrates that in this setting, U-net is consistently able to detect and label accurately. The faster R-CNN performs well for volumes of individual bugs, but for both types of mixtures, it fails to detect many bugs. The nnDetection has poor performance for the volumes of individual bugs, but performs well for the two synthetic mixtures.

\input{images/train_det_examples}

\clearpage

\bibliographystyle{splncs04}
\bibliography{ref}

\end{document}

%% file: images/illustration1_v2.tex
\hspace*{-2.1em}
\begin{tikzpicture}
    \node[anchor=north west, inner sep=0] (image) at (0,0) {\includegraphics[width=\linewidth]{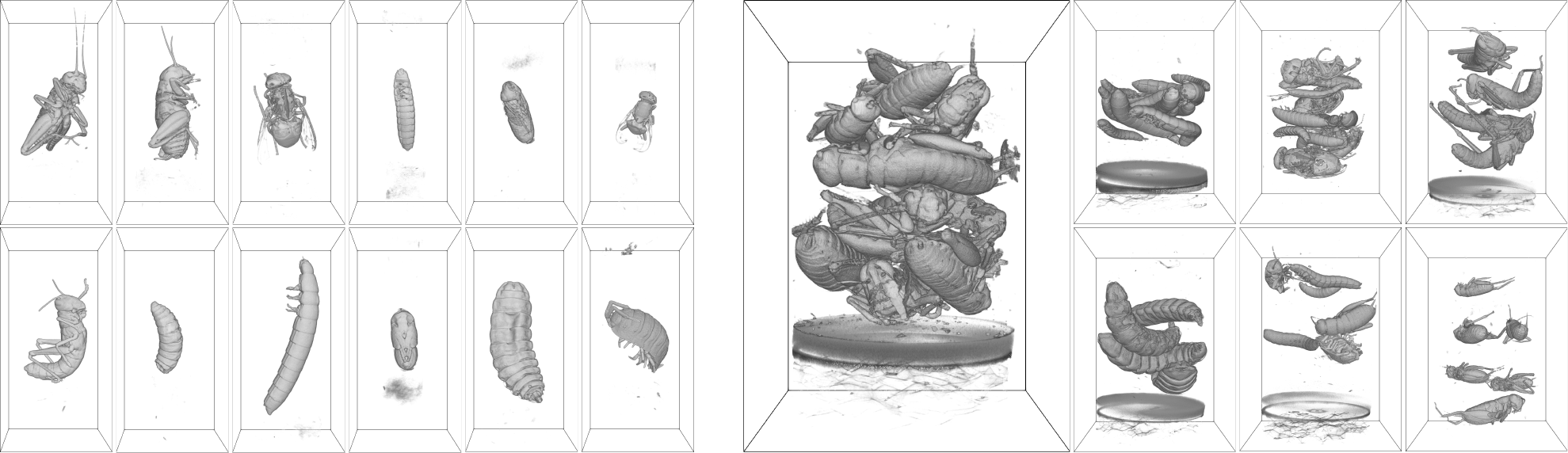}};
    
    \pgfmathsetmacro\offset{0.22}
    \pgfmathsetmacro\boxSpacing{0.45pt}
    \def\pointSize{3pt}

    \foreach \i/\text/\subtitle/\count in {
        0/AC/Brown\\cricket/724,
        1/BC/Black\\cricket/761,
        2/BF/Blow\\fly/733,
        3/BL/Beetle\\larva/773,
        4/BP/Blow fly\\pupa/748,
        5/CF/Curly-\\wing fly/756
    } {
        \draw (\offset + \boxSpacing*\i,-0.12) node[fill=\text, inner sep = 1.5, text=white, anchor=center, font=\scriptsize] at +(\offset + \boxSpacing*\i,0) {\text};
        \draw (\offset + \boxSpacing*\i, -1.1) node[text=black, anchor=north, yshift=-2, font=\scriptsize, scale=0.9] at +(\offset + \boxSpacing*\i,0) {\parbox[t]{2cm}{\centering\vphantom{gl}\subtitle}};
        \draw (\offset + \boxSpacing*\i, -0.12) node[text=black, anchor=north, yshift=-2, font=\scriptsize, scale=0.95] at +(\offset + \boxSpacing*\i,0) {\parbox[t]{2cm}{\centering\count}};
    }
    
    \foreach \i/\text/\subtitle/\count in {
        0/GH/Grass-\\hopper/713,
        1/MA/Maggot/758,
        2/ML/Meal-\\worm/737,
        3/PP/Fly\\pupa/765,
        4/SL/Soldier\\fly larva/740,
        5/WO/Wood-\\lice/946
    } {
        \draw (\offset + \boxSpacing*\i, -1.90) node[fill=\text, inner sep = 1.5, text=white, anchor=center, font=\scriptsize] at +(\offset + \boxSpacing*\i,0) {\text};
        \draw (\offset + \boxSpacing*\i, -2.88) node[text=black, anchor=north, yshift=-2, font=\scriptsize, scale=0.9] at +(\offset + \boxSpacing*\i,0) {\parbox[t]{2cm}{\centering\vphantom{gl}\subtitle}};
        \draw (\offset + \boxSpacing*\i, -1.90) node[text=black, anchor=north, yshift=-2, font=\scriptsize] at +(\offset + \boxSpacing*\i,0) {\parbox[t]{2cm}{\centering\count}};
    }

    \draw[fill=BC, draw=white, line width=0.5pt] (7.2, -0.65) circle (\pointSize);
    \fill[BC] (6.55, -0.0) rectangle (7.05, -0.3);
    \draw[-, line width=1.25pt, color=BC] (7.0, -0.25) -- (7.2, -0.65);
    \draw (6.8, -0.15) node[text=white, anchor=center, font=\scriptsize]  {BC};

    \draw[fill=BC, draw=white, line width=0.5pt] (6.6, -1.0) circle (\pointSize);
    \fill[BC] (5.9, -0.35) rectangle (6.4, -0.65);
    \draw[-, line width=1.25pt, color=BC] (6.35, -0.6) -- (6.6, -1.0);
    \draw (6.15, -0.5) node[text=white, anchor=center, font=\scriptsize] {BC};
    
    \draw[fill=AC, draw=white, line width=0.5pt] (7.45, -0.8) circle (\pointSize);
    \fill[AC] (7.8, -0.15) rectangle (8.3, -0.45);
    \draw[-, line width=1.25pt, color=AC] (7.85, -0.4) -- (7.45, -0.8);
    \draw (8.05, -0.3) node[text=white, anchor=center, font=\scriptsize]  {AC};

    \draw[fill=BC, draw=white, line width=0.5pt] (7.1, -1.3) circle (\pointSize);
    \fill[BC] (7.7, -0.75) rectangle (8.2, -1.05);
    \draw[-, line width=1.25pt, color=BC] (7.75, -1.02) -- (7.1, -1.3);
    \draw (7.95, -0.9) node[text=white, anchor=center, font=\scriptsize]  {BC};

    \draw[fill=AC, draw=white, line width=0.5pt] (7.2, -1.6) circle (\pointSize);
    \fill[AC] (5.65, -1.05) rectangle (6.15, -1.35);
    \draw[-, line width=1.25pt, color=AC] (6.1, -1.32) -- (7.2, -1.6);
    \draw (5.9, -1.2) node[text=white, anchor=center, font=\scriptsize]  {AC};

    \draw[fill=BC, draw=white, line width=0.5pt] (7.5, -1.65) circle (\pointSize);
    \fill[BC] (7.8, -1.3) rectangle (8.3, -1.6);
    \draw[-, line width=1.25pt, color=BC] (7.85, -1.58) -- (7.5, -1.65);
    \draw (8.05, -1.45) node[text=white, anchor=center, font=\scriptsize]  {BC};

    \draw[fill=AC, draw=white, line width=0.5pt] (6.7, -1.75) circle (\pointSize);
    \fill[AC] (5.7, -1.4) rectangle (6.2, -1.7);
    \draw[-, line width=1.25pt, color=AC] (6.15, -1.68) -- (6.7, -1.75);
    \draw (5.95, -1.55) node[text=white, anchor=center, font=\scriptsize]  {AC};
    
    \draw[fill=SL, draw=white, line width=0.5pt] (6.4, -2.0) circle (\pointSize);
    \fill[SL] (5.85, -2.35) rectangle (6.35, -2.65);
    \draw[-, line width=1.25pt, color=SL] (6.33, -2.37) -- (6.4, -2.0);
    \draw (6.1, -2.5) node[text=white, anchor=center, font=\scriptsize]  {SL};

    \draw[fill=AC, draw=white, line width=0.5pt] (7.4, -2.1) circle (\pointSize);
    \fill[AC] (7.75, -2.1) rectangle (8.25, -2.4);
    \draw[-, line width=1.25pt, color=AC] (7.8, -2.12) -- (7.4, -2.1);
    \draw (8.0, -2.25) node[text=white, anchor=center, font=\scriptsize]  {AC};
    
    \draw[fill=GH, draw=white, line width=0.5pt] (6.9, -2.25) circle (\pointSize);
    \fill[GH] (6.7, -2.6) rectangle (7.2, -2.9);
    \draw[-, line width=1.25pt, color=GH] (6.72, -2.62) -- (6.9, -2.25);
    \draw (6.95, -2.75) node[text=white, anchor=center, font=\scriptsize]  {GH};

    \draw[fill=ML, draw=white, line width=0.5pt] (8.66, -0.84) circle (0.6*\pointSize);
    \draw[fill=BL, draw=white, line width=0.5pt] (8.75, -1.05) circle (0.6*\pointSize);
    \draw[fill=BL, draw=white, line width=0.5pt] (8.785, -0.721) circle (0.6*\pointSize);
    \draw[fill=ML, draw=white, line width=0.5pt] ( 8.855 , -0.959 ) circle (0.6*\pointSize);
    \draw[fill=ML, draw=white, line width=0.5pt] ( 8.96  , -0.784 ) circle (0.6*\pointSize);
    \draw[fill=ML, draw=white, line width=0.5pt] ( 9.1   , -0.945 ) circle (0.6*\pointSize);
    \draw[fill=ML, draw=white, line width=0.5pt] ( 9.128 , -0.826 ) circle (0.6*\pointSize);
    \draw[fill=ML, draw=white, line width=0.5pt] ( 9.135 , -0.714 ) circle (0.6*\pointSize);
    \draw[fill=BL, draw=white, line width=0.5pt] ( 9.31  , -0.63  ) circle (0.6*\pointSize);
    
    \draw[fill=GH, draw=white, line width=0.5pt] (10.29  , -0.56  ) circle (0.6*\pointSize);
    \draw[fill=BL, draw=white, line width=0.5pt] (10.36  , -0.63  ) circle (0.6*\pointSize);
    \draw[fill=ML, draw=white, line width=0.5pt] (10.255 , -0.735 ) circle (0.6*\pointSize);
    \draw[fill=WO, draw=white, line width=0.5pt] (10.325 , -0.798 ) circle (0.6*\pointSize);
    \draw[fill=ML, draw=white, line width=0.5pt] (10.115 , -0.924 ) circle (0.6*\pointSize);
    \draw[fill=BL, draw=white, line width=0.5pt] (10.36  , -0.945 ) circle (0.6*\pointSize);
    \draw[fill=SL, draw=white, line width=0.5pt] (10.465 , -0.959 ) circle (0.6*\pointSize);
    \draw[fill=ML, draw=white, line width=0.5pt] (10.08  , -1.015 ) circle (0.6*\pointSize);
    \draw[fill=ML, draw=white, line width=0.5pt] (10.255 , -1.085 ) circle (0.6*\pointSize);
    \draw[fill=WO, draw=white, line width=0.5pt] (10.43  , -1.12  ) circle (0.6*\pointSize);
    \draw[fill=BF, draw=white, line width=0.5pt] (10.22  , -1.2425) circle (0.6*\pointSize);
    
    \draw[fill=GH, draw=white, line width=0.5pt] (11.585 , -0.42  ) circle (0.6*\pointSize);
    \draw[fill=GH, draw=white, line width=0.5pt] (11.725 , -0.735 ) circle (0.6*\pointSize);
    \draw[fill=GH, draw=white, line width=0.5pt] (11.69  , -0.98  ) circle (0.6*\pointSize);
    \draw[fill=GH, draw=white, line width=0.5pt] (11.76  , -1.12  ) circle (0.6*\pointSize);
    
    \draw[fill=SL, draw=white, line width=0.5pt] ( 9.065 , -2.415 ) circle (0.6*\pointSize);
    \draw[fill=SL, draw=white, line width=0.5pt] ( 8.89  , -2.625 ) circle (0.6*\pointSize);
    \draw[fill=SL, draw=white, line width=0.5pt] ( 8.995 , -2.8   ) circle (0.6*\pointSize);
    \draw[fill=SL, draw=white, line width=0.5pt] ( 9.1875, -2.975 ) circle (0.6*\pointSize);
    
    \draw[fill=AC, draw=white, line width=0.5pt] ( 9.94  , -2.17  ) circle (0.6*\pointSize);
    \draw[fill=ML, draw=white, line width=0.5pt] (10.29  , -2.1875) circle (0.6*\pointSize);
    \draw[fill=BL, draw=white, line width=0.5pt] (10.3075, -2.296 ) circle (0.6*\pointSize);
    \draw[fill=AC, draw=white, line width=0.5pt] (10.5   , -2.52  ) circle (0.6*\pointSize);
    \draw[fill=ML, draw=white, line width=0.5pt] (10.08  , -2.66  ) circle (0.6*\pointSize);
    \draw[fill=WO, draw=white, line width=0.5pt] (10.43  , -2.73  ) circle (0.6*\pointSize);
    
    \draw[fill=BC, draw=white, line width=0.5pt] (11.515 , -2.247 ) circle (0.6*\pointSize);
    \draw[fill=BC, draw=white, line width=0.5pt] (11.83  , -2.59  ) circle (0.6*\pointSize);
    \draw[fill=BC, draw=white, line width=0.5pt] (11.48  , -2.604 ) circle (0.6*\pointSize);
    \draw[fill=BC, draw=white, line width=0.5pt] (11.445 , -2.9225) circle (0.6*\pointSize);
    \draw[fill=BC, draw=white, line width=0.5pt] (11.795 , -2.996 ) circle (0.6*\pointSize);
    \draw[fill=AC, draw=white, line width=0.5pt] (11.69  , -3.185 ) circle (0.6*\pointSize);
    
    \draw (2.75, -4.1) node[text=black, anchor=center, font=\scriptsize]  {\parbox[t]{0.375\linewidth}{(a) 9154 volumes of individual bugs with class labels.}};
    \draw (8.95, -4.1) node[text=black, anchor=center, font=\scriptsize]  {\parbox[t]{0.45\linewidth}{(b) 388 volumes of bug mixtures with center point and class labels.}};
\end{tikzpicture}

%% file: images/mixes_summary.tex
\begin{figure}[t]
    \centering
    \begin{tikzpicture}
        \node at (-0.25\linewidth, 0.0125\linewidth) {\includegraphics[width=0.42\linewidth]{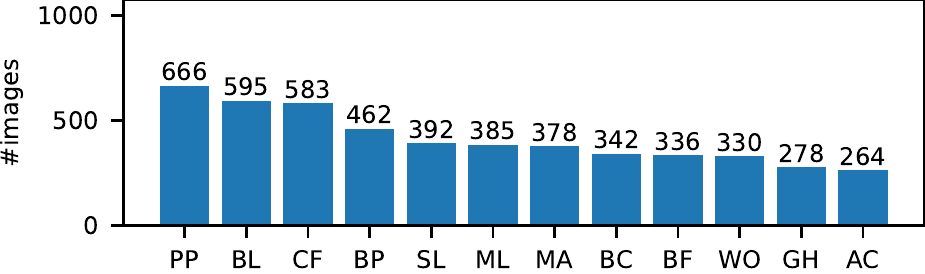}};
        \node at (0.125\linewidth, 0) {\includegraphics[width=0.23\linewidth]{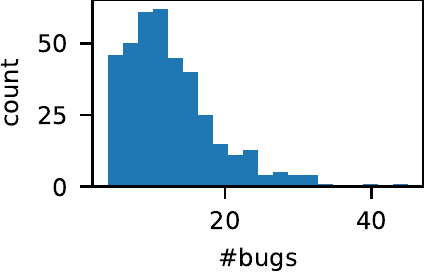}};
        \node at (0.375\linewidth, 0) {\includegraphics[width=0.23\linewidth]{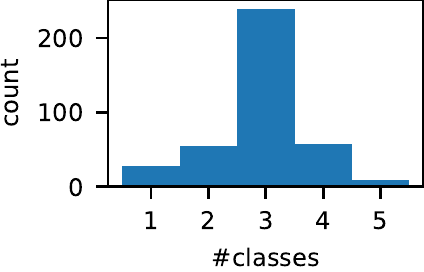}};  
    \node[text=black, font=\scriptsize] at (-0.25\linewidth, -1.25) {\parbox[t]{0.4\linewidth}{(a) Class distribution for bugs from  mixture volumes.}};
    \node[text=black, font=\scriptsize] at (0.25\linewidth, -1.25) {\parbox[t]{0.45\linewidth}{(b) Distribution of the number of bugs per volume and the number of classes per volume.}};
    \end{tikzpicture}
    \caption{Summary of volumes from scans of bug mixtures.}
    \label{fig:mixes_summary}
\end{figure}

%% file: images/illustration3.tex
\begin{tikzpicture}

\def\ss{0.7}

\node at (0, 0) {\includegraphics[width=0.8\linewidth]{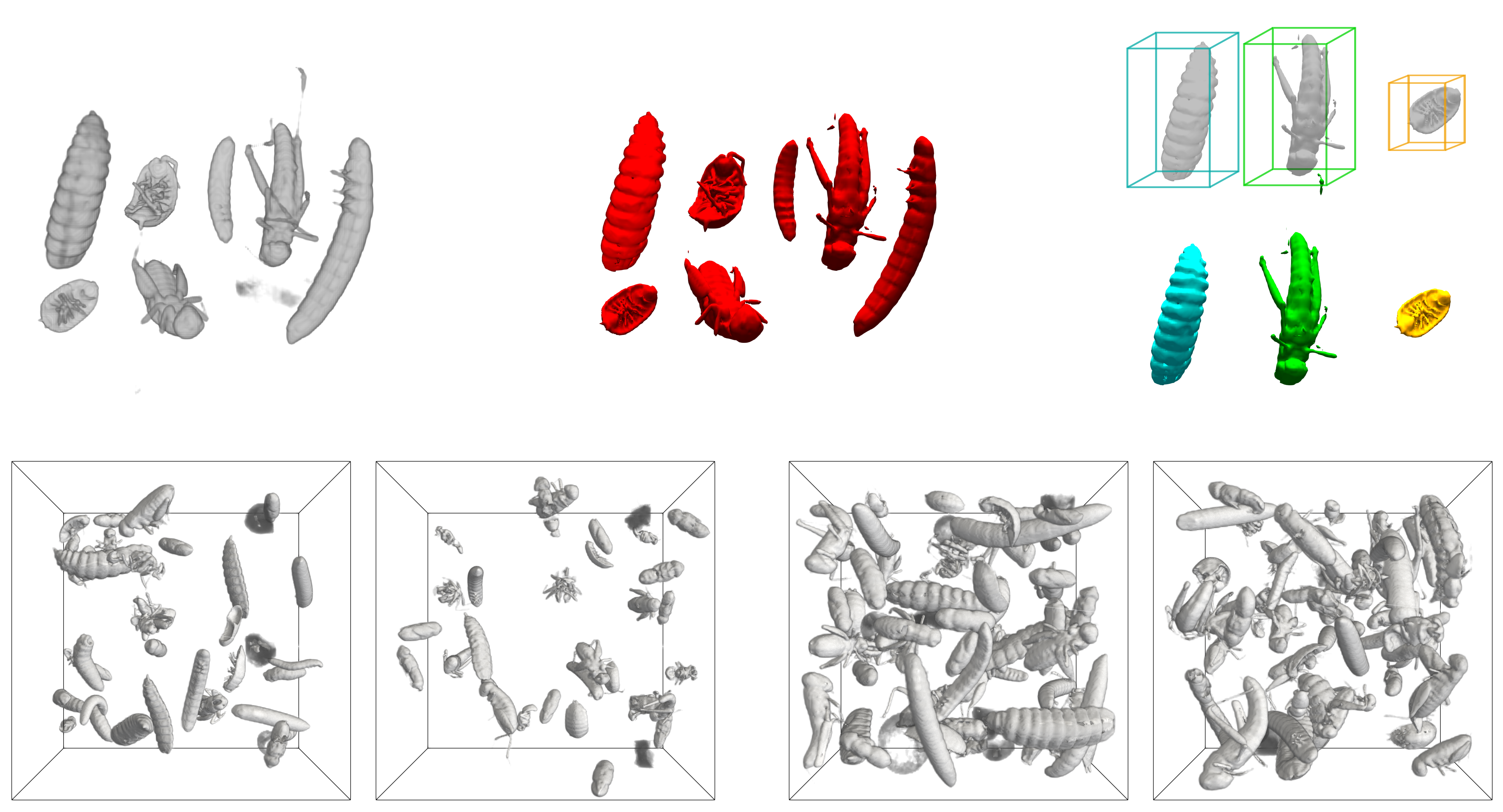}};
\node[font=\scriptsize, anchor=west] at (-6, 2.3) {Automated annotation pipeline};
\node[font=\scriptsize, anchor=west] at (-4.8, 2.1) {\scalebox{\ss}{Volumes of individual bugs}};
\node[font=\scriptsize] at (0.2, 2.1) {\scalebox{\ss}{Surfaces of bugs}};
\node[font=\scriptsize] at (-1.7, 1.4) {\scalebox{\ss}{Thresholding}};
\draw[->] (-2.3, 1.2) -- (-1.1, 1.2);

\node[font=\scriptsize] at (1.8, 1.7) {\scalebox{\ss}{\parbox{1.5cm} {\centering Automatic \\ annotation }}};
\draw[->] (1.3, 1.1) -- (2.3, 1.4);
\draw[->] (1.3, 0.9) -- (2.3, 0.6);

\node[font=\scriptsize] at (3.5, 1.3) {\scalebox{\ss}{Bounding boxes}};
\node[font=\scriptsize] at (3.5, 0.05) {\scalebox{\ss}{Segmentations}};

\node[font=\scriptsize, anchor=west] at (-6, 0.0) {Synthetically generated mixtures};
\node[font=\scriptsize, anchor=west] at (-4.8, -0.2) {\scalebox{\ss}{Less crowded}};
\node[font=\scriptsize] at (1, -0.2) {\scalebox{\ss}{More crowded}};

\end{tikzpicture}	

%% file: images/match_illustration.tex
\begin{figure}
    \centering
    \begin{tikzpicture}
    \newlength{\pf}\setlength{\pf}{0.25\linewidth}
        \node at (-1.05\pf, 0) {\includegraphics[angle=90, trim=0.9cm 0 0.9cm 0, clip, width=\pf]{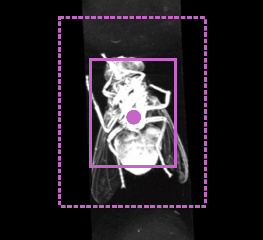}};
        \node at (0, 0) {\includegraphics[angle=90, trim=0.9cm 0 0.9cm 0, clip, width=\pf]{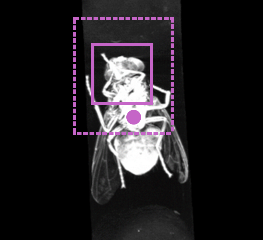}};
        \node at (1.05\pf, 0) {\includegraphics[angle=90, trim=0.9cm 0 0.9cm 0, clip, width=\pf]{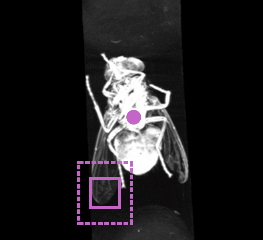}};  
        \node[text=black, font=\scriptsize] at (-1.05\pf, -1.5) {(a)};
        \node[text=black, font=\scriptsize] at (0, -1.5) {(b)};
        \node[text=black, font=\scriptsize] at (1.05\pf, -1.5) {(c)};
    \end{tikzpicture}
    \caption{Matching criterion for bounding box to center point. The circle shows the manually annotated center point, the predicted bounding box is shown with the full line, and the inflated bounding box is shown with the dashed line. In (a), the match is accepted by the predicted bounding box, in (b), the match is also accepted due to the inflated bounding box, whereas in (c), there is no match. Note that this illustration is 2D, but the detection is done in 3D. }
    \label{fig:match-illustration}
\end{figure}

%% file: tables/bugnist_detection_results.tex
\begin{table*}[ht!]
    \centering
    \scriptsize
    \setlength{\tabcolsep}{2pt}
    \begingroup
    \renewcommand*{\arraystretch}{1.1}
    \def\catspace{6pt}
    \caption{Detection results on mixed scans. We show mean and std. for each metric computed over the evaluation images. Higher is better. Without class info refers to not considering the bug label and with class info expects the bug type to be classified as well as being detected.}
    \begin{tabularx}{\linewidth}{lXrrr@{\hskip \catspace}rrr}
    \toprule
    & & \multicolumn{3}{c}{Without class info.} & \multicolumn{3}{c}{With class info.} \\
    Model & - Trained on & F1-Score & Precision & Recall & F1-Score & Precision & Recall \\
    \midrule
    U-Net 
    & - Single bugs & $0.59 \!\pm\!0.18$ & $0.68 \!\pm\!0.17$ & $0.60 \!\pm\!0.25$ & $0.11 \!\pm\!0.16$ & $0.13 \!\pm\!0.19$ & $0.11 \!\pm\!0.16$ \\
    \cite{cciccek20163d} 
    & - Synth. mixes & $0.60 \!\pm\!0.12$ & $0.52 \!\pm\!0.17$ & $0.82 \!\pm\!0.19$ & $0.11 \!\pm\!0.08$ & $0.09 \!\pm\!0.08$ & $0.14 \!\pm\!0.10$ \\
    & - Crowded s.m. & $0.54 \!\pm\!0.13$ & $0.39 \!\pm\!0.14$ & $0.97 \!\pm\!0.07$ & $0.10 \!\pm\!0.08$ & $0.07 \!\pm\!0.06$ & $0.18 \!\pm\!0.14$ \\
    \midrule
    Faster 
    & - Single bugs & $0.56 \!\pm\!0.20$ & $1.00 \!\pm\!0.04$ & $0.42 \!\pm\!0.21$ & $0.11 \!\pm\!0.10$ & $0.21 \!\pm\!0.19$ & $0.08 \!\pm\!0.08$ \\
    R-CNN
    & - Synth. mixes & $0.68 \!\pm\!0.16$ & $0.63 \!\pm\!0.21$ & $0.83 \!\pm\!0.22$ & $0.16 \!\pm\!0.14$ & $0.15 \!\pm\!0.14$ & $0.19 \!\pm\!0.18$ \\
    \cite{ren2015faster} 
    & - Crowded s.m. & $0.27 \!\pm\!0.17$ & $0.81 \!\pm\!0.24$ & $0.18 \!\pm\!0.16$ & $0.06 \!\pm\!0.09$ & $0.19 \!\pm\!0.30$ & $0.04 \!\pm\!0.06$ \\
    \midrule
    nn-
    & - Single bugs & $0.03 \!\pm\!0.02$ & $0.02 \!\pm\!0.01$ & $0.91 \!\pm\!0.15$ & $0.00 \!\pm\!0.01$ & $0.00 \!\pm\!0.00$ & $0.09 \!\pm\!0.16$ \\
    Detection
    & - Synth. mixes & $0.50 \!\pm\!0.14$ & $0.41 \!\pm\!0.17$ & $0.80 \!\pm\!0.22$ & $0.10 \!\pm\!0.06$ & $0.08 \!\pm\!0.06$ & $0.16 \!\pm\!0.11$ \\
    \cite{baumgartner2021nndetection} 
    & - Crowded s.m. & $0.60 \!\pm\!0.19$ & $0.77 \!\pm\!0.23$ & $0.58 \!\pm\!0.27$ & $0.11 \!\pm\!0.12$ & $0.15 \!\pm\!0.16$ & $0.11 \!\pm\!0.12$ \\ 
    \bottomrule
    \end{tabularx}
    \label{tab:det_results}
    \endgroup
\end{table*}

%% file: images/mix_detection.tex

\begin{figure*}[ht]
    \centering
    \setlength{\tabcolsep}{2pt}
    \scriptsize
    \def\figcapspace{0.155\linewidth}
    \begin{tabular}{lc}
    & 
    \makebox[0.465\linewidth][c]{Best without class info} 
    \makebox[0.465\linewidth][c]{Best with class info} 
    \\
    &
    \makebox[\figcapspace][c]{F1-score}%
    \makebox[\figcapspace][c]{Precision}%
    \makebox[\figcapspace][c]{Recall}%
    \makebox[\figcapspace][c]{F1-score}%
    \makebox[\figcapspace][c]{Precision}%
    \makebox[\figcapspace][c]{Recall}%
    \\
    \rotatebox[origin=c]{90}{U-Net \cite{cciccek20163d}} & \raisebox{-.5\height}{\includegraphics[width=0.95\linewidth]{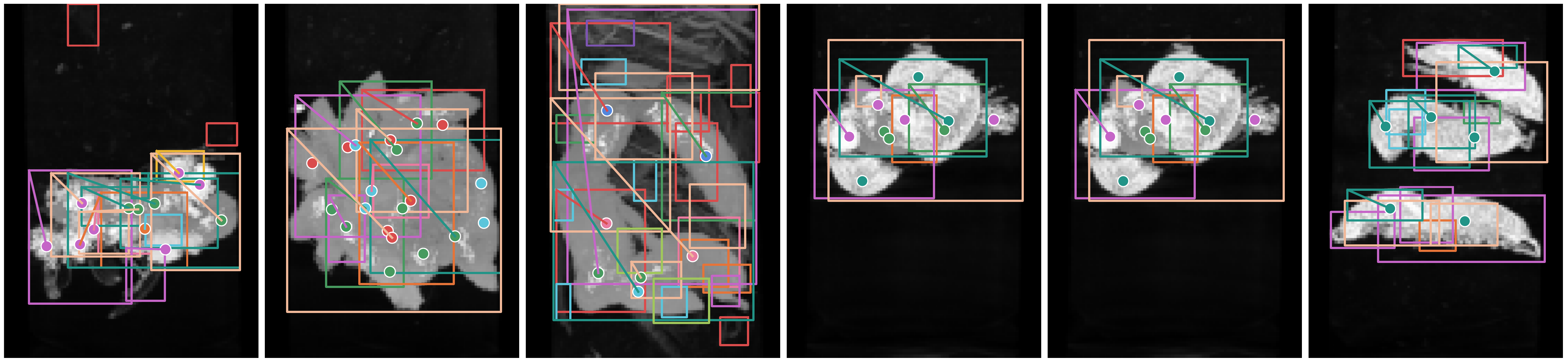}} \\
    \rotatebox[origin=c]{90}{Faster R-CNN \cite{ren2015faster}}& \raisebox{-.5\height}{\includegraphics[width=0.95\linewidth]{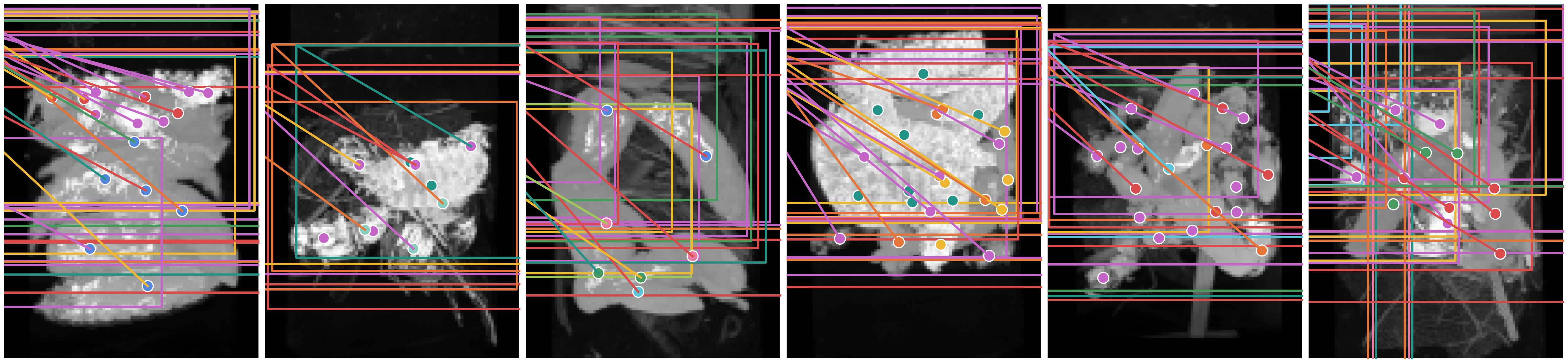}} \\
    \rotatebox[origin=c]{90}{nnDetection \cite{baumgartner2021nndetection}}& \raisebox{-.5\height}{\includegraphics[width=0.95\linewidth]{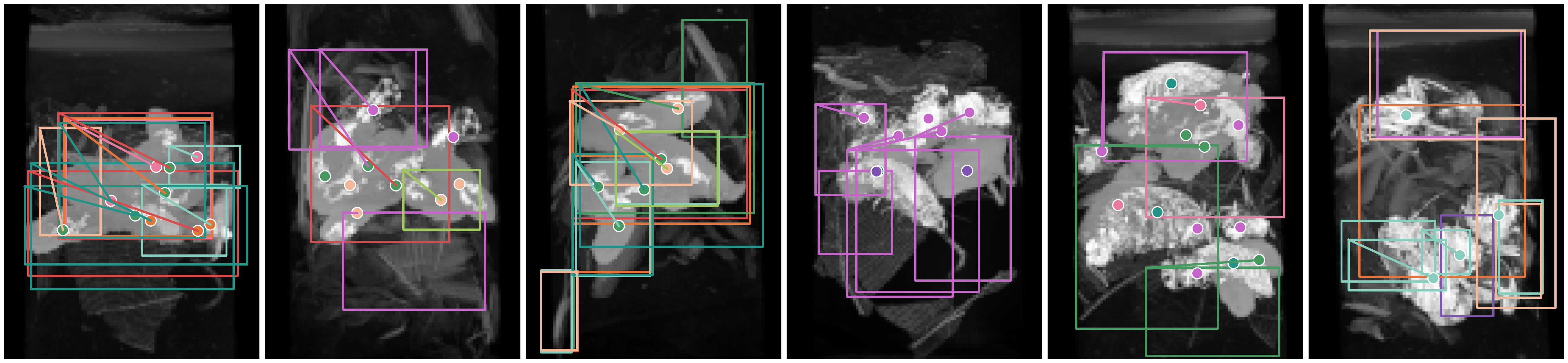}}
    \end{tabular}
    \caption{Detection results on real mixed scans. The images are maximum-intensity projections of the volumes. Center points are marked with circles, detections are marked with bounding boxes, and color indicates the bug class. A line from a point to the upper left corner of a bounding box shows a match. If the center point and the bounding box have the same color, then the detection and classification are correct. The shown images are the volumes with the highest scores for the different models. For U-Net the results when trained on synthetic mixes, for Faster R-CNN when trained on synthetic mixes and nnDetection when trained on crowded synthetic mixes.}
    \label{fig:mix_detection}
\end{figure*}

%% file: images/metric_hists.tex
\begin{figure}[ht!]
    \centering
    \begin{tikzpicture}
        \node at (-0.33\linewidth, 0) {\includegraphics[width=0.3\linewidth]{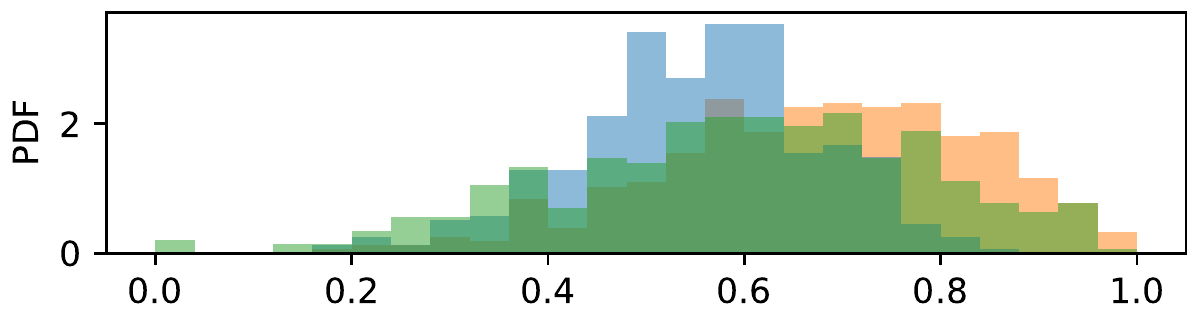}};
        \node at (0, 0) {\includegraphics[width=0.3\linewidth]{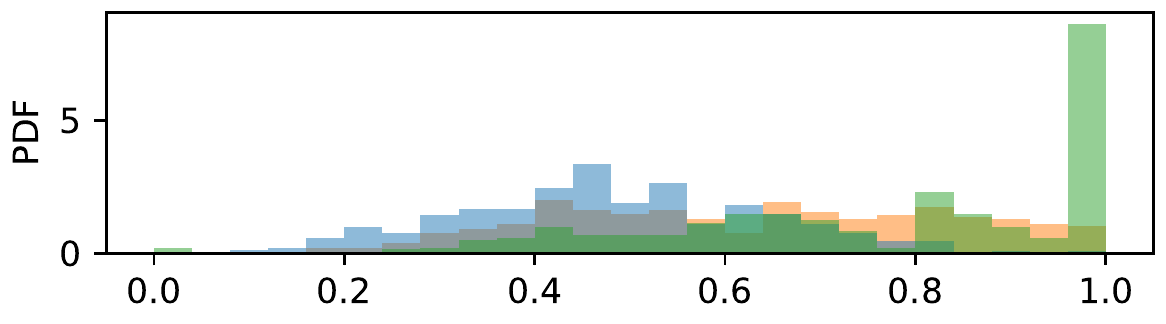}};
        \node at (0.33\linewidth, 0) {\includegraphics[width=0.3\linewidth]{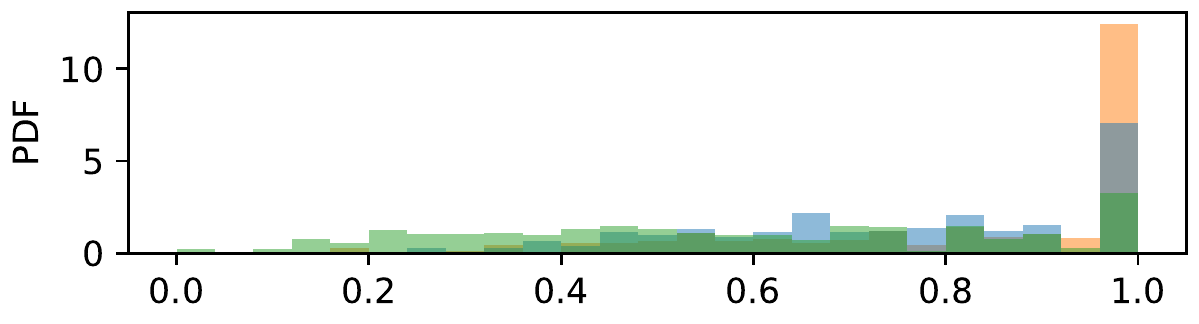}};
        \node at (-0.33\linewidth, -0.65) {\scriptsize F1-score};
        \node at (0, -0.65) {\scriptsize Precision};
        \node at (0.33\linewidth, -0.65) {\scriptsize Recall};    
        \node at (0, -1){\includegraphics[width=0.4\linewidth]{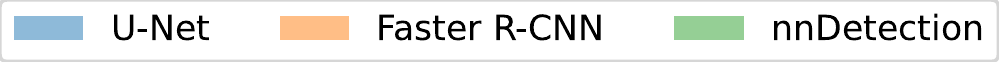}};
    \end{tikzpicture}
    \caption{Distributions of performance scores for the three detection methods.}
    \label{fig:metric_hists}
\end{figure}

%% file: tables/train_task_results.tex
\begin{table}[ht!]
    \centering
    \scriptsize
    \setlength{\tabcolsep}{3pt}
    \caption{Detection results on the training tasks obtained during training on the single bugs and the two synthetic mixtures. For U-Net and Faster R-CNN performance is evaluated on a validation split. Since nnDetection is an ensemble created from multiple train-validation splits, we evaluate on all images in the training task.  We show mean average precision (mAP) between predicted and ground truth bounding boxes. Higher is better.}
    \begin{tabularx}{\linewidth}{lXrr}
    \toprule
    Model & - Trained on & mAP & mAP@50 \\
    \midrule
    U-Net \cite{cciccek20163d} & - Single bugs & 0.72 & 0.72 \\
    & - Synthetic mixes & 0.64 & 0.77 \\
    & - Crowded synth. mixes & 0.75 & 0.86 \\
    \midrule
    Faster R-CNN \cite{ren2015faster} & - Single bugs & 0.70 & 0.70 \\
    & - Synthetic mixes & 0.02 & 0.02 \\
    & - Crowded synth. mixes & 0.05 & 0.05 \\
    \midrule
    nnDetection \cite{baumgartner2021nndetection} & - Single bugs & 0.10 & 0.10 \\
    & - Synthetic mixes & 0.59 & 0.70 \\
    & - Crowded synth. mixes & 0.49 & 0.66\\
    \bottomrule
    \end{tabularx}
    \label{tab:train_det_results}
\end{table}

%% file: images/illustration3_512_64.tex
\def\xs{0.08\linewidth}
\def\sc{1.025}

\begin{tikzpicture}
	
	\foreach \a/\i in {AC/1, BC/2, BF/3, BL/4, BP/5, CF/6, GH/7, MA/8, ML/9, PP/10, SL/11, WO/12}
	{
		\node at (\i*\sc, -2.85){\includegraphics[width=\xs]{images/sup_illustration/x512/max_\a.png}};
		\node at (\i*\sc, -4.6){\includegraphics[width=\xs]{images/sup_illustration/x512/crossmax_\a.png}};
		\node at (\i*\sc, -6.85){\includegraphics[width=\xs]{images/sup_illustration/x256/max_\a.png}};
		\node at (\i*\sc, -8.6){\includegraphics[width=\xs]{images/sup_illustration/x256/crossmax_\a.png}};
		\node at (\i*\sc, -10.85){\includegraphics[width=\xs]{images/sup_illustration/x128/max_\a.png}};
		\node at (\i*\sc, -12.6){\includegraphics[width=\xs]{images/sup_illustration/x128/crossmax_\a.png}};
		\node at (\i*\sc, -14.85){\includegraphics[width=\xs]{images/sup_illustration/x64/max_\a.png}};
		\node at (\i*\sc, -16.6){\includegraphics[width=\xs]{images/sup_illustration/x64/crossmax_\a.png}};
	}
 \node at (6.5*\sc, -3.25 + 1.7) {Volume size $512 \times 256 \times 256$};
 \node at (6.5*\sc, -7.25 + 1.7) {Volume size $256 \times 128 \times 128$};
 \node at (6.5*\sc, -11.25 + 1.7) {Volume size $128 \times 64 \times 64$};
 \node at (6.5*\sc, -15.25 + 1.7) {Volume size $64 \times 32 \times 32$};
\end{tikzpicture}	

%% file: images/illustration4.tex
\def\xs{0.14\linewidth}
\def\sc{1.8}

\begin{tikzpicture}
	\foreach \a/\i in {10_000/1, 10_001/2, 6_009/3, 10_003/4, 10_004/5, 10_005/6, 10_013/7}
	{
		\node at (\i*\sc, -3){\includegraphics[width=\xs]{images/mixtures/x512/max_mixture_\a.png}};
		\node at (\i*\sc, -3 - 2.2){\includegraphics[width=\xs]{images/mixtures/x512/crossmax_mixture_\a.png}};
		\node at (\i*\sc, -8.1){\includegraphics[width=\xs]{images/mixtures/x128/max_mixture_\a.png}};
		\node at (\i*\sc, -8.1 - 2.2){\includegraphics[width=\xs]{images/mixtures/x128/crossmax_mixture_\a.png}};
		\node at (\i*\sc, -13.2){\includegraphics[width=\xs]{images/mixtures/x64/max_mixture_\a.png}};
		\node at (\i*\sc, -13.2 - 2.2){\includegraphics[width=\xs]{images/mixtures/x64/crossmax_mixture_\a.png}};
        }
   \node at (4*\sc, -3 + 1.5) {Volume size $512 \times 370 \times 370$};
   \node at (4*\sc, -8.1 + 1.5) {Volume size $128 \times 92 \times 92$};
   \node at (4*\sc, -13.2 + 1.5) {Volume size $64 \times 47 \times 47$};
 \end{tikzpicture}	

%% file: images/train_det_examples.tex
\begin{figure*}[ht]
    \centering
    \setlength{\tabcolsep}{0pt}
    \renewcommand{\arraystretch}{0}
    \small
    \newlength{\indh}\setlength{\indh}{0.16\linewidth}
    \newlength{\fix}\setlength{\fix}{\dimexpr -.16\linewidth + 2pt\relax}
    \begin{tabular}{l@{\hskip 4pt}c@{\hskip 2pt}c@{\hskip 2pt}c}
    \rotatebox[origin=c]{90}{\makebox[\dimexpr 0.32\linewidth + 2pt\relax][c]{U-Net \cite{cciccek20163d}}} & 
    \begin{tabular}{cccc}
        \includegraphics[height=\indh]{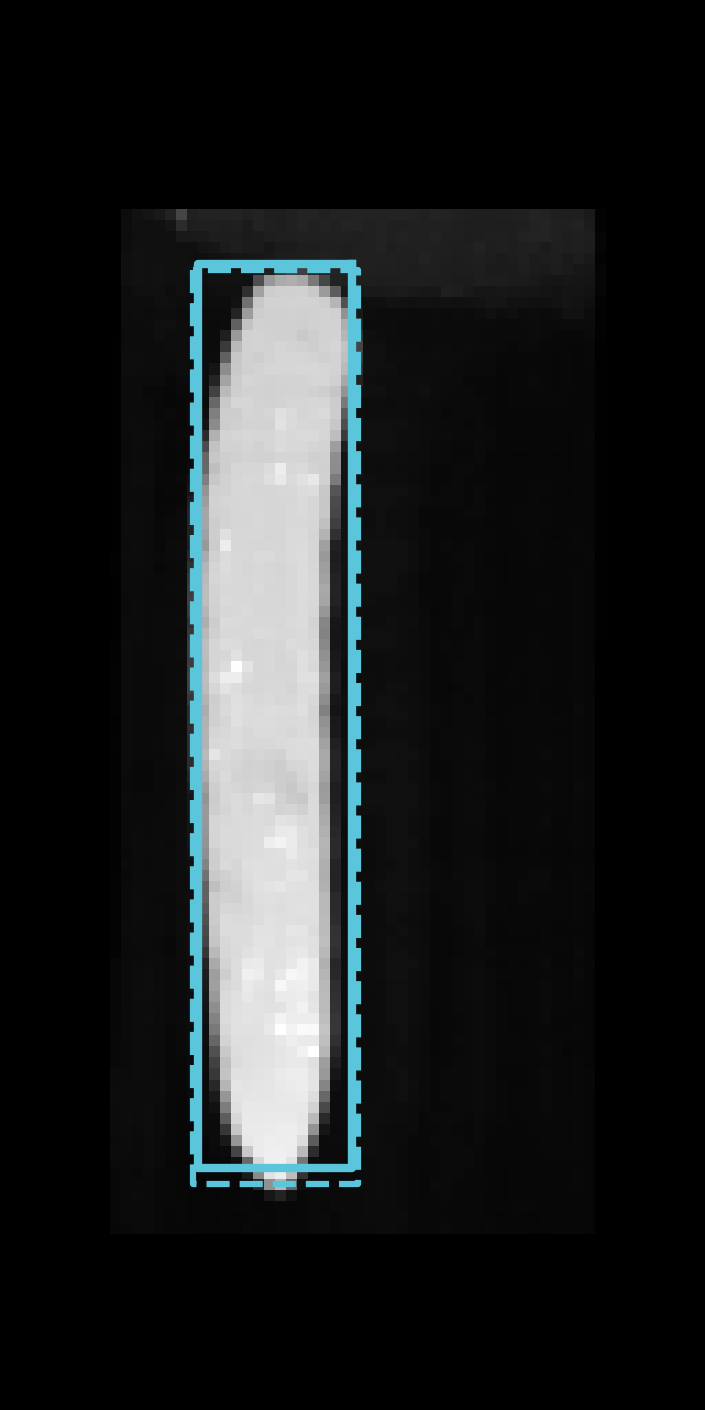} & 
        \includegraphics[height=\indh]{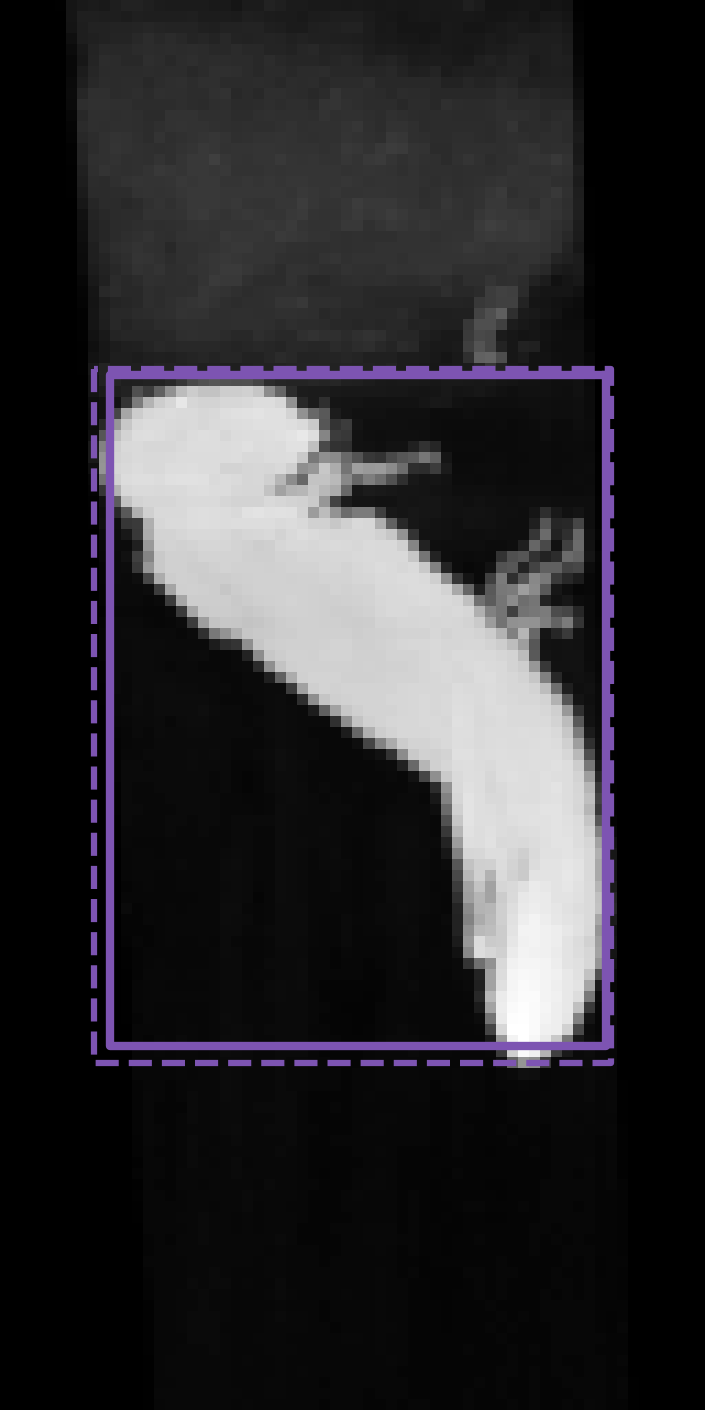} & 
        \includegraphics[height=\indh]{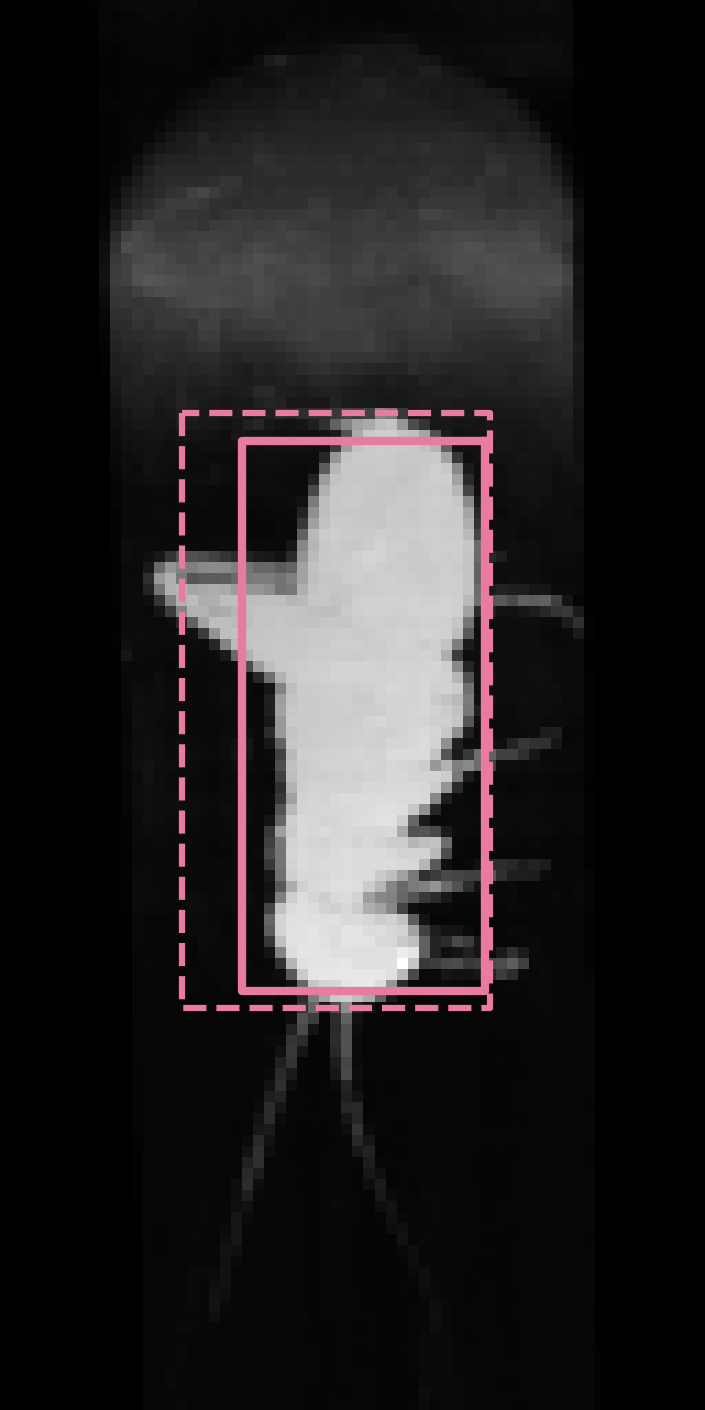} &
        \includegraphics[height=\indh]{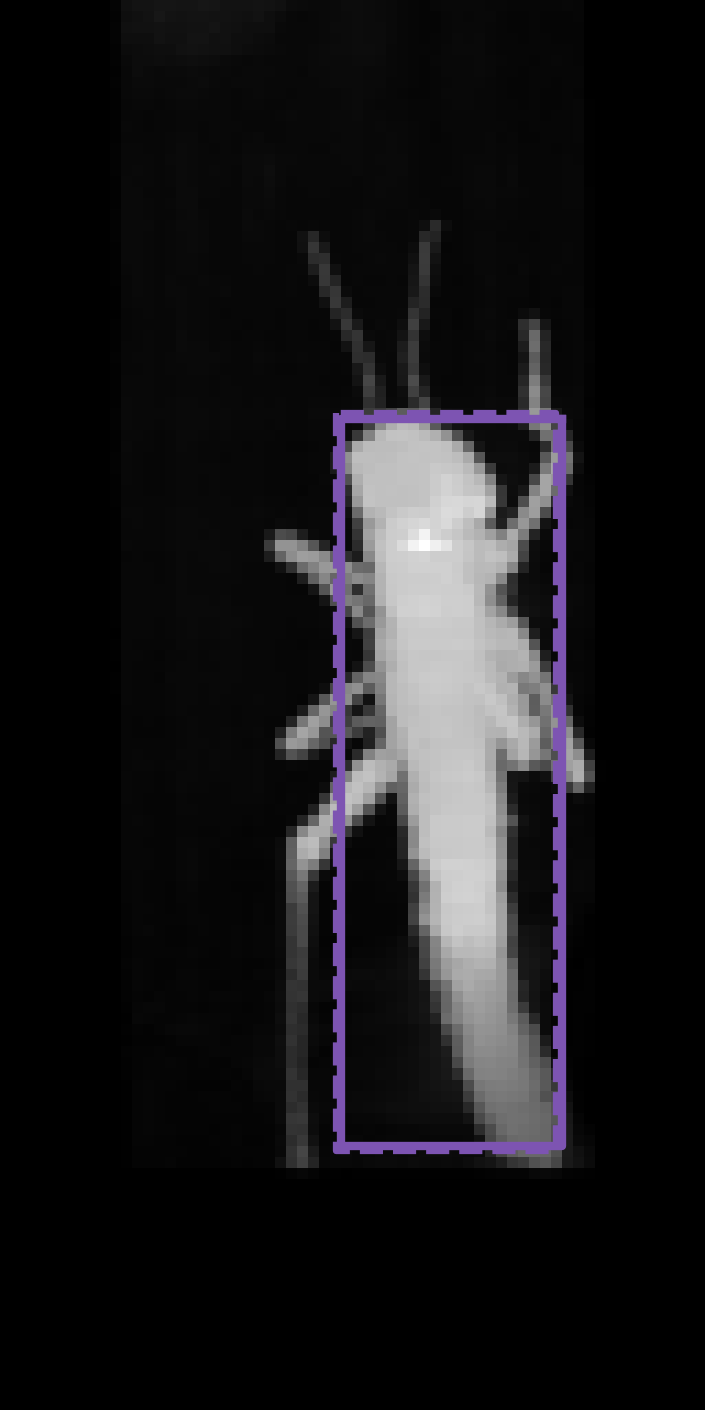} \\
        \includegraphics[height=\indh]{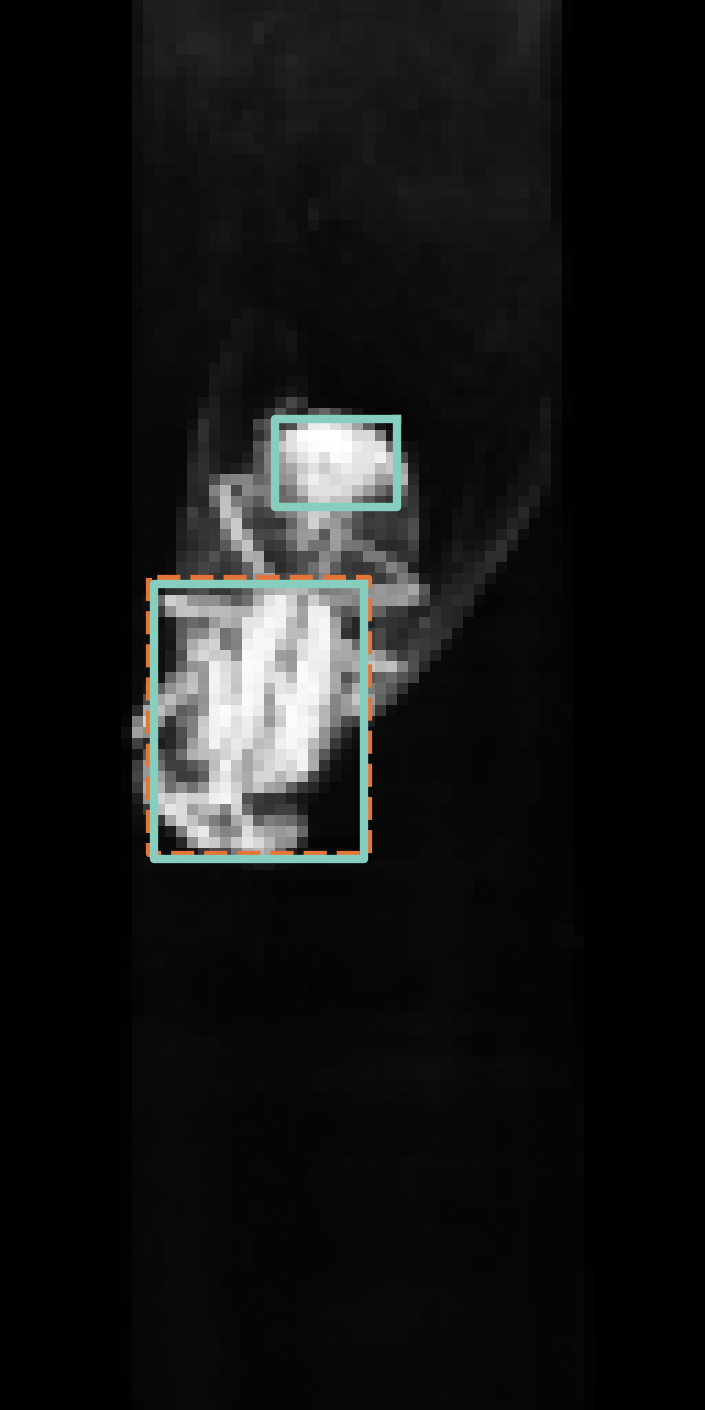} & 
        \includegraphics[height=\indh]{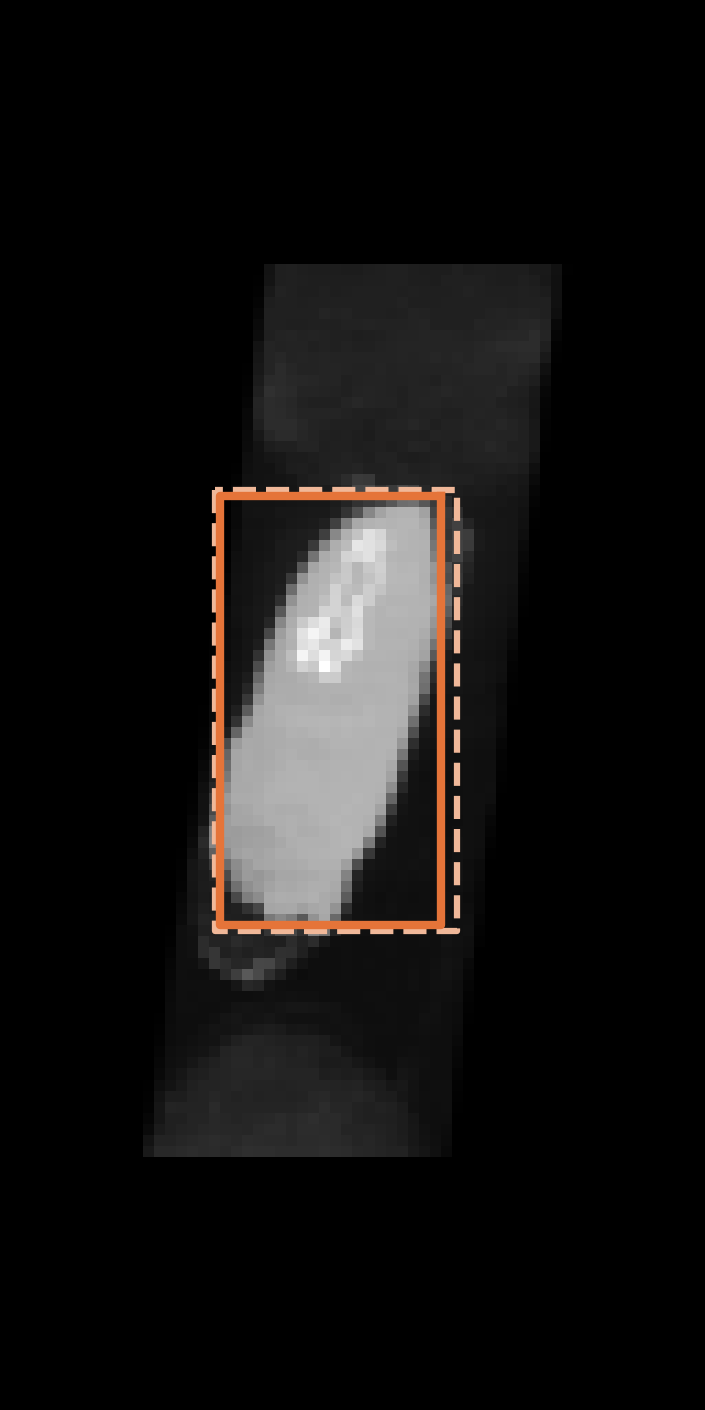} & 
        \includegraphics[height=\indh]{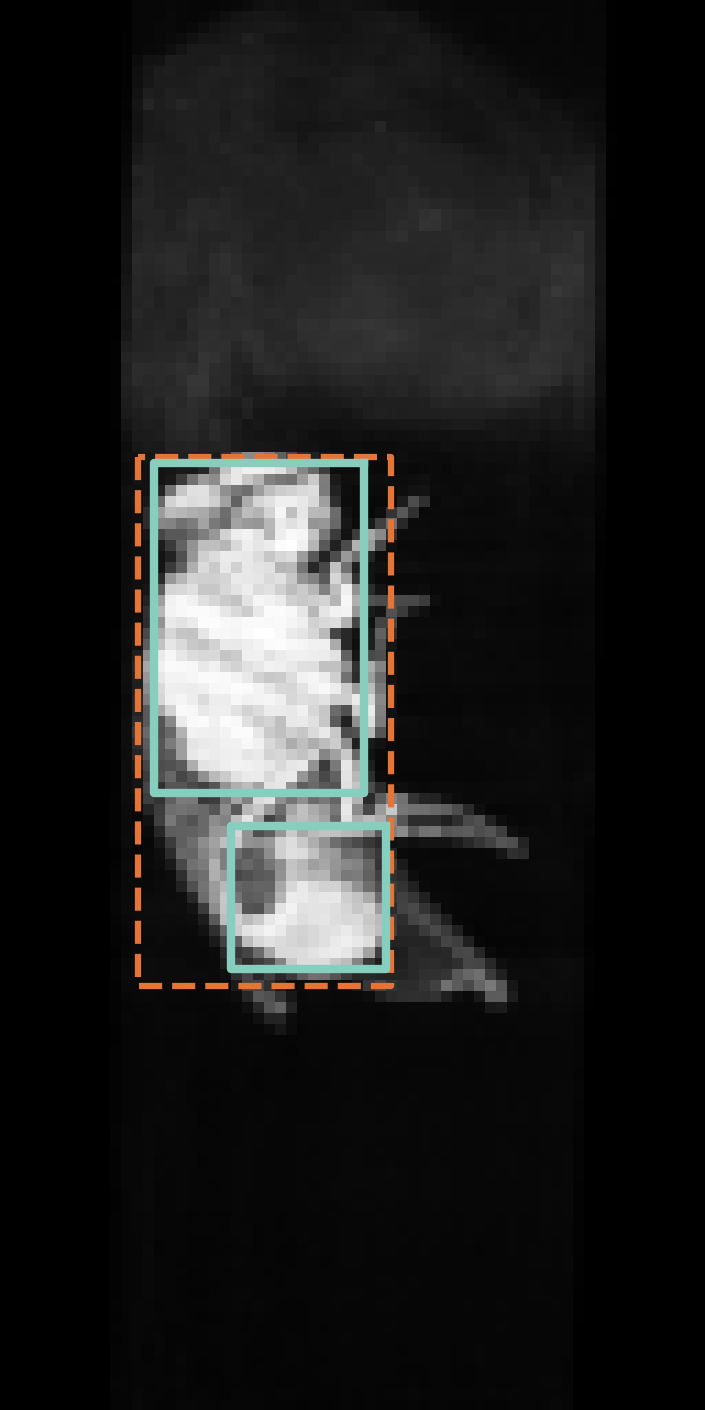} & 
        \includegraphics[height=\indh]{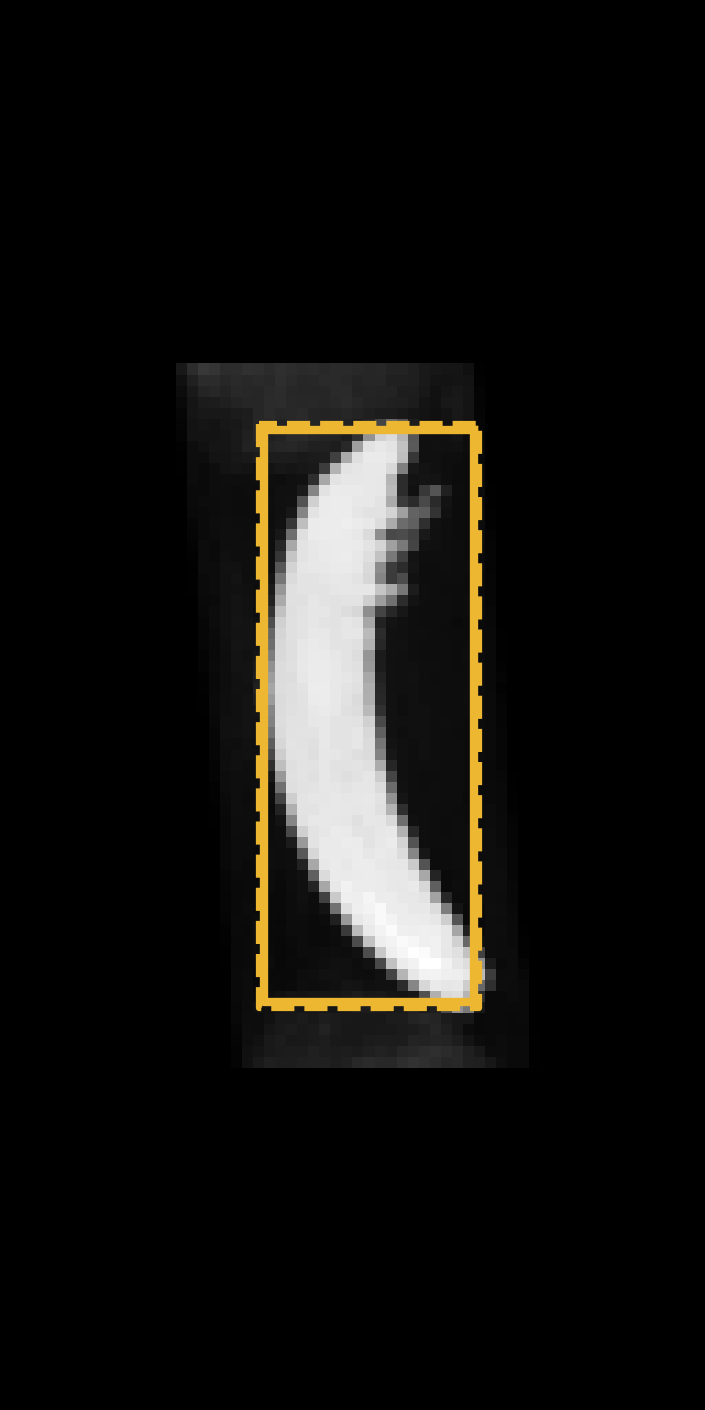}
    \end{tabular} & 
    \raisebox{\fix}{\includegraphics[width=0.32\linewidth]{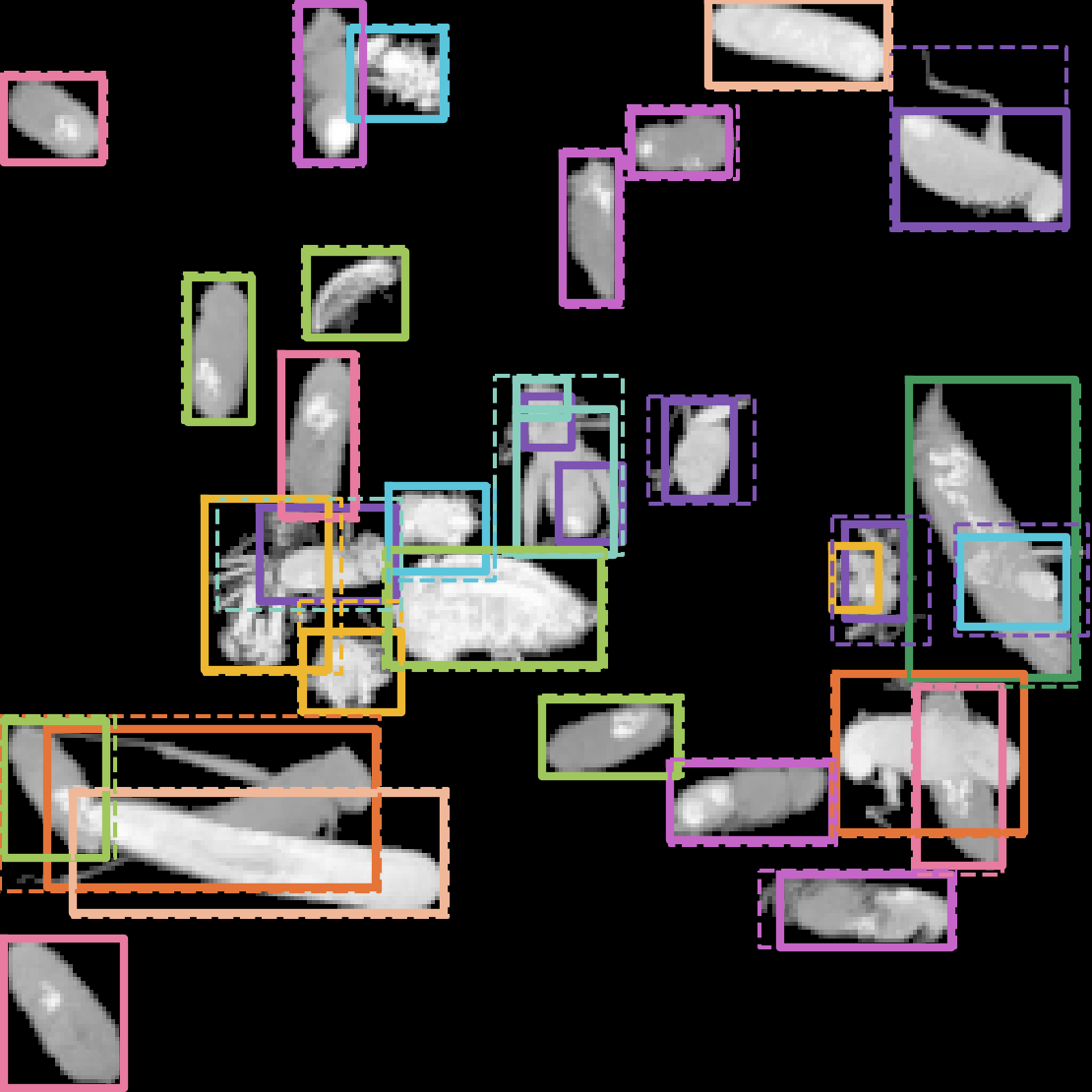}} & 
    \raisebox{\fix}{\includegraphics[width=0.32\linewidth]{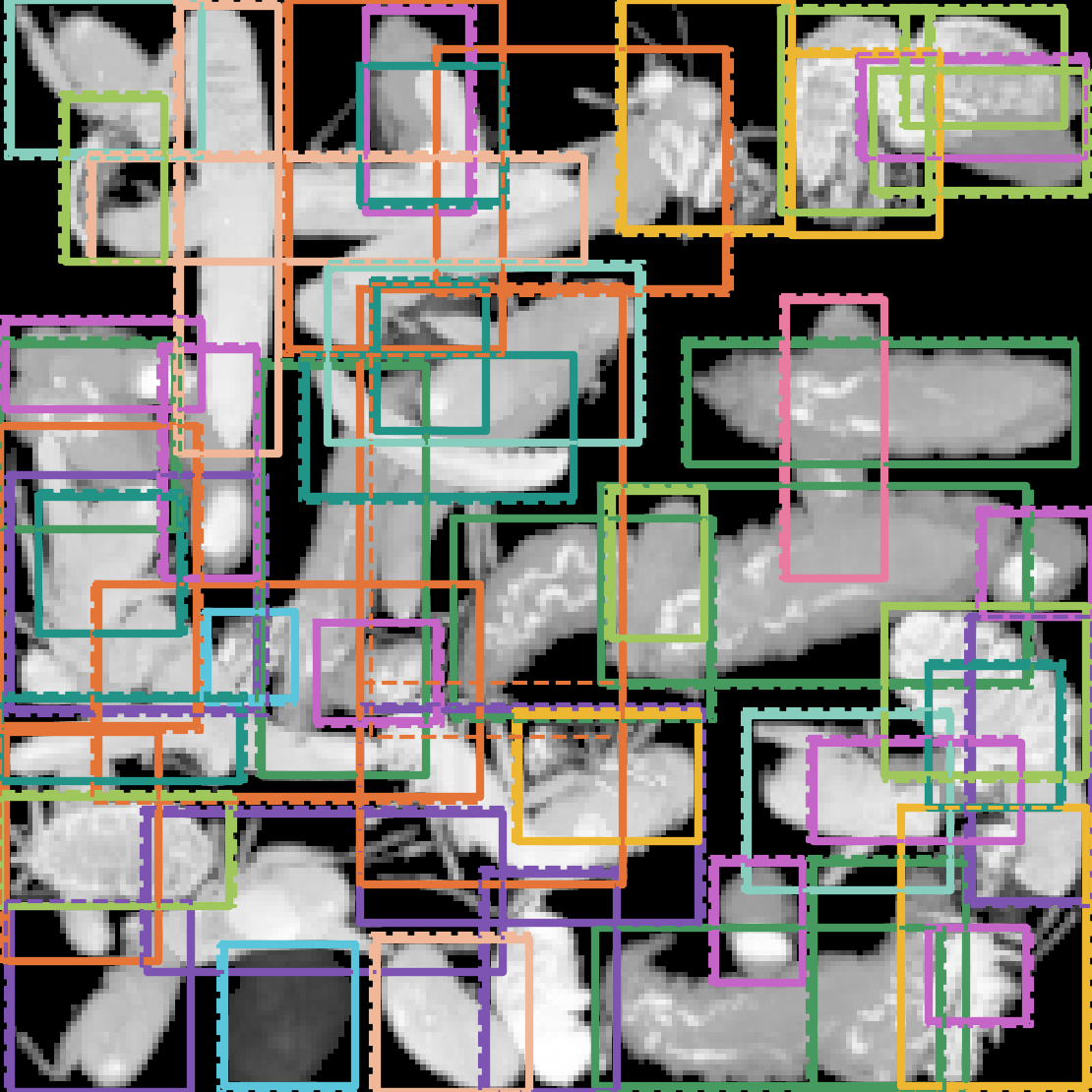}} \\ 
    \rotatebox[origin=c]{90}{\makebox[\dimexpr 0.32\linewidth + 2pt\relax][c]{Faster R-CNN \cite{ren2015faster}}} & 
    \begin{tabular}{cccc}
        \includegraphics[height=\indh]{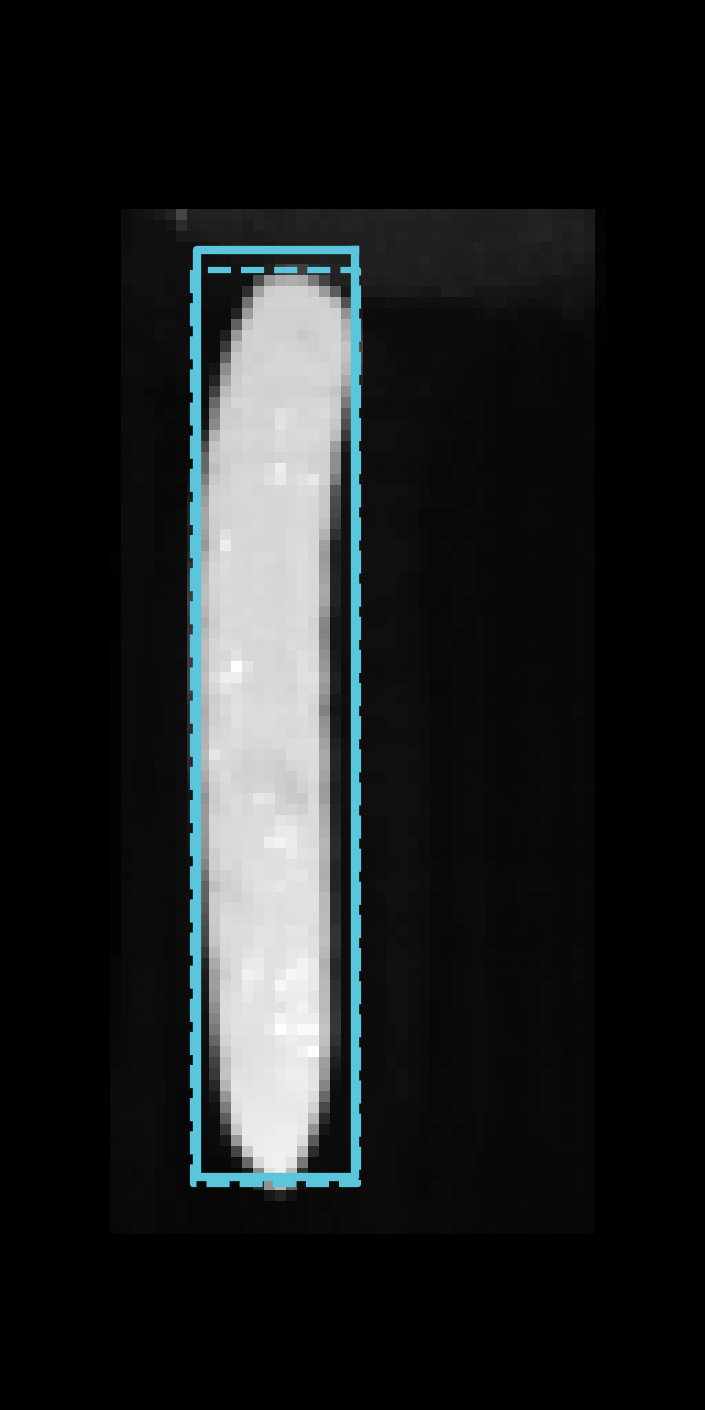} & 
        \includegraphics[height=\indh]{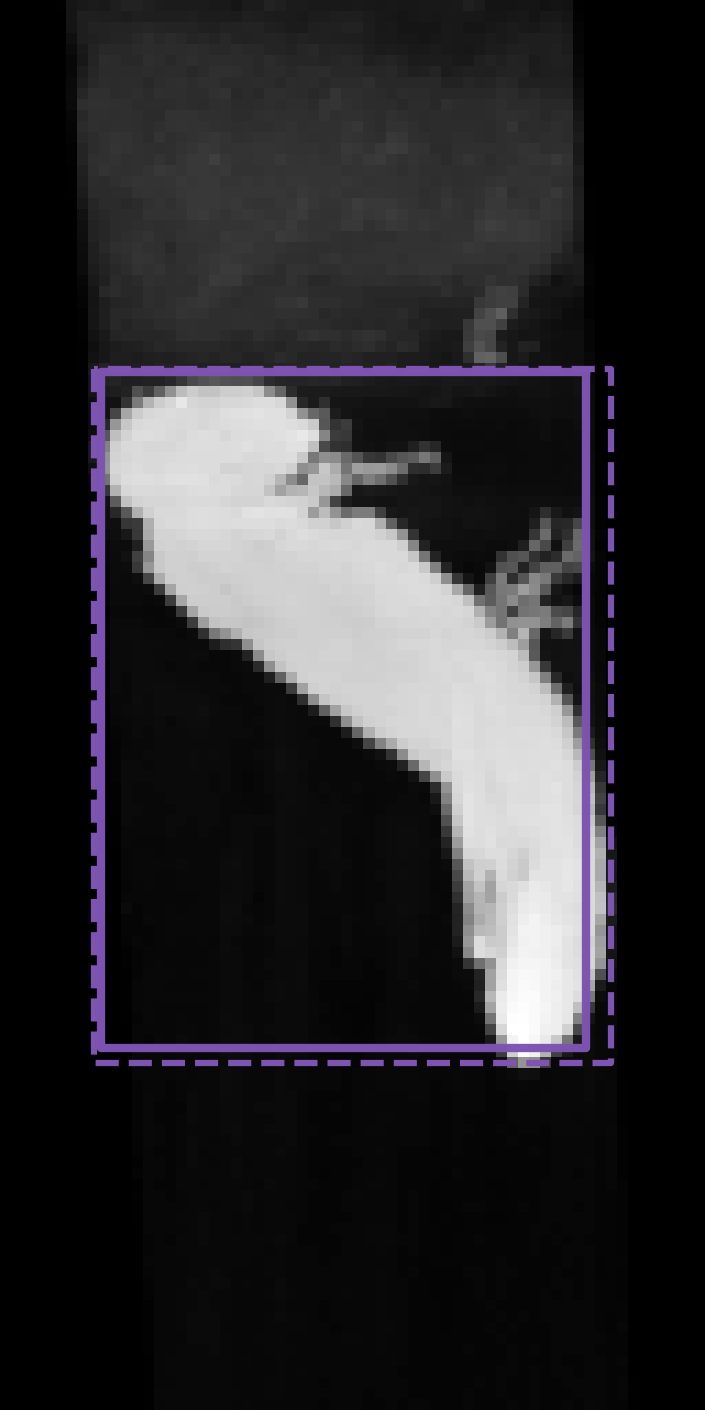} & 
        \includegraphics[height=\indh]{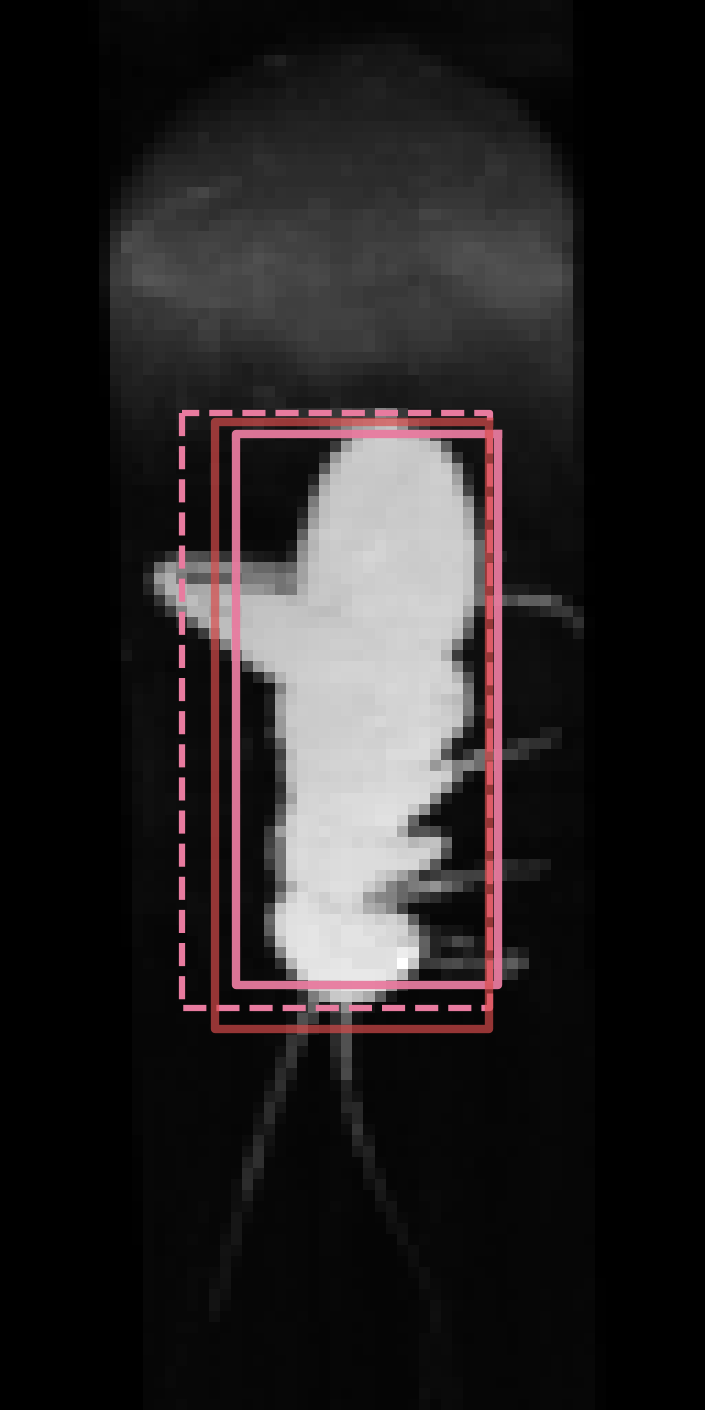} &
        \includegraphics[height=\indh]{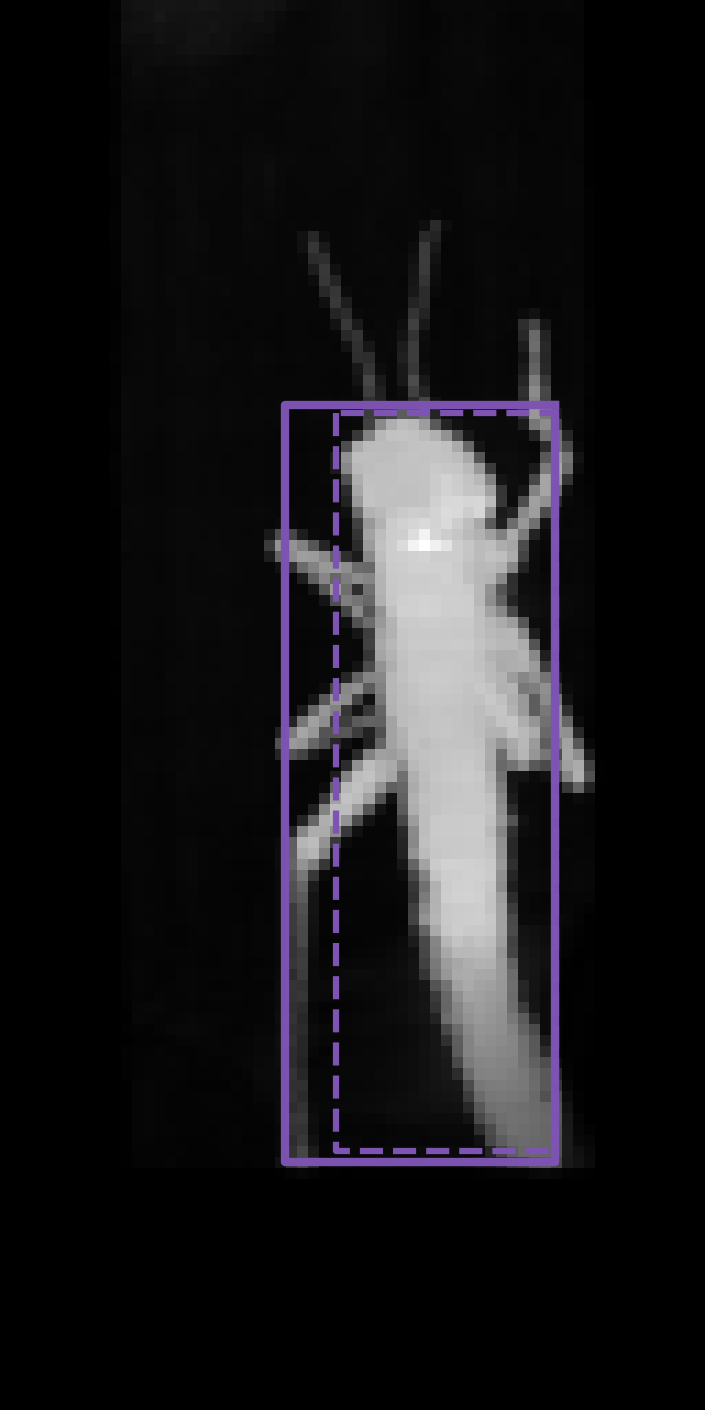} \\
        \includegraphics[height=\indh]{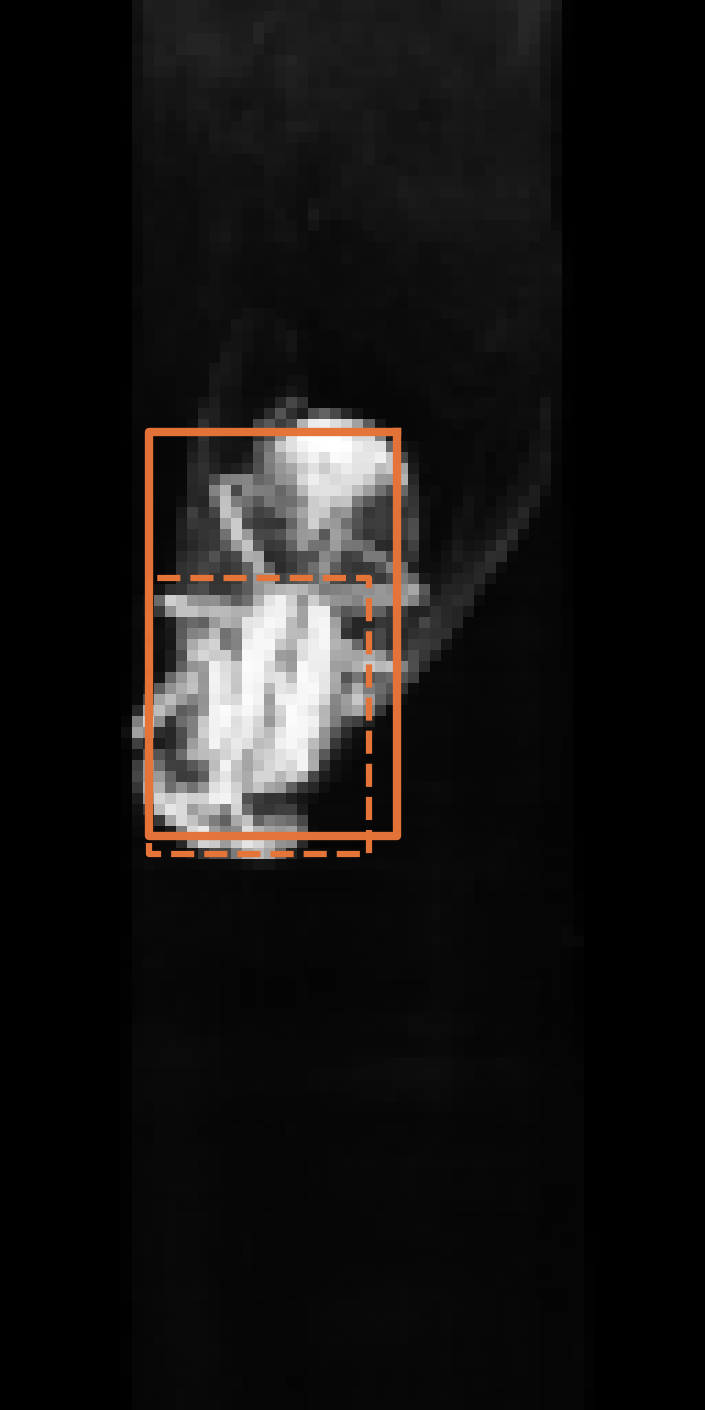} & 
        \includegraphics[height=\indh]{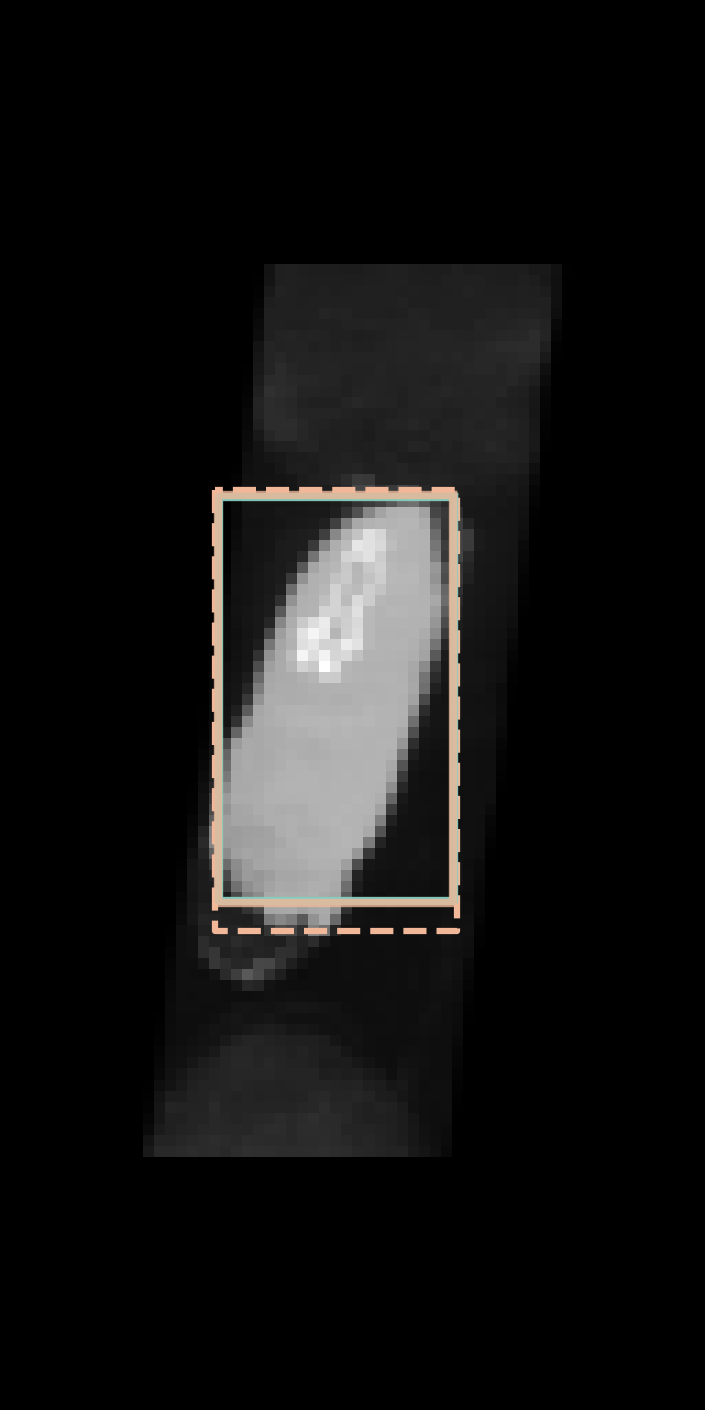} & 
        \includegraphics[height=\indh]{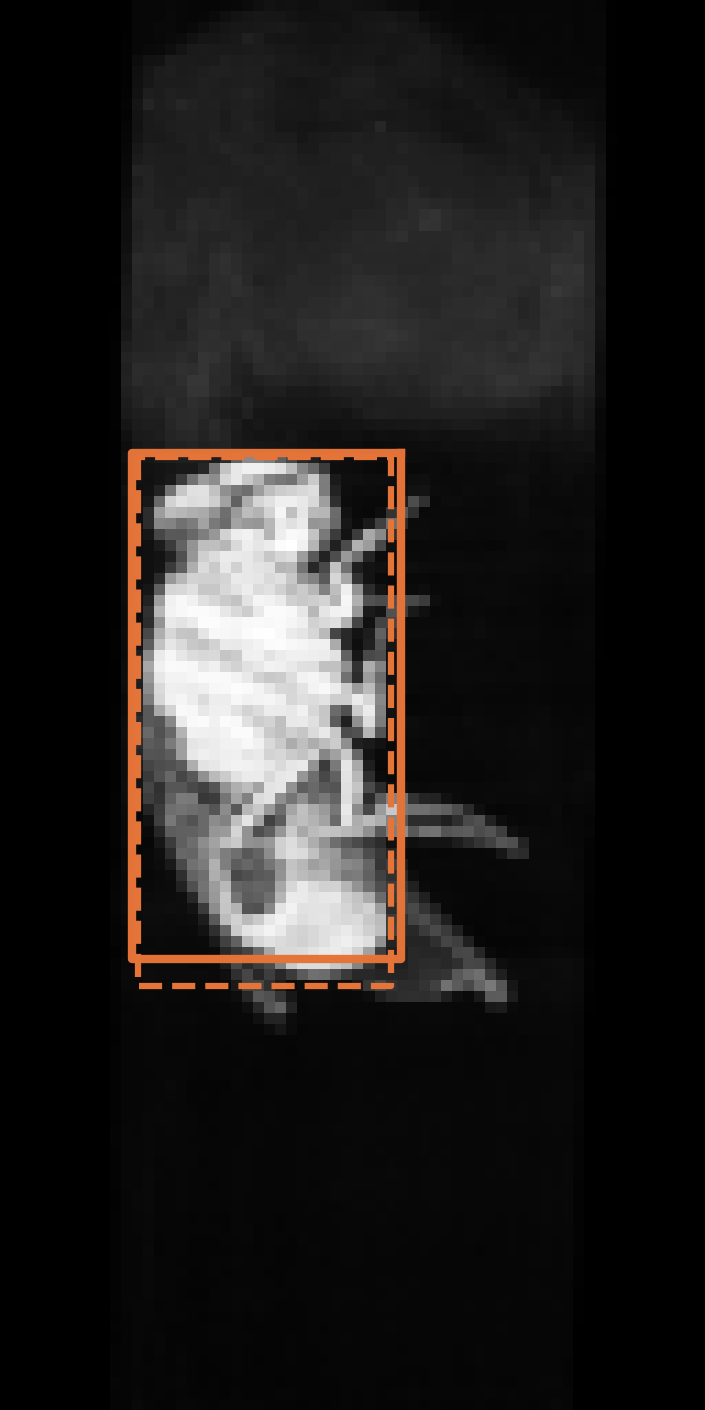} & 
        \includegraphics[height=\indh]{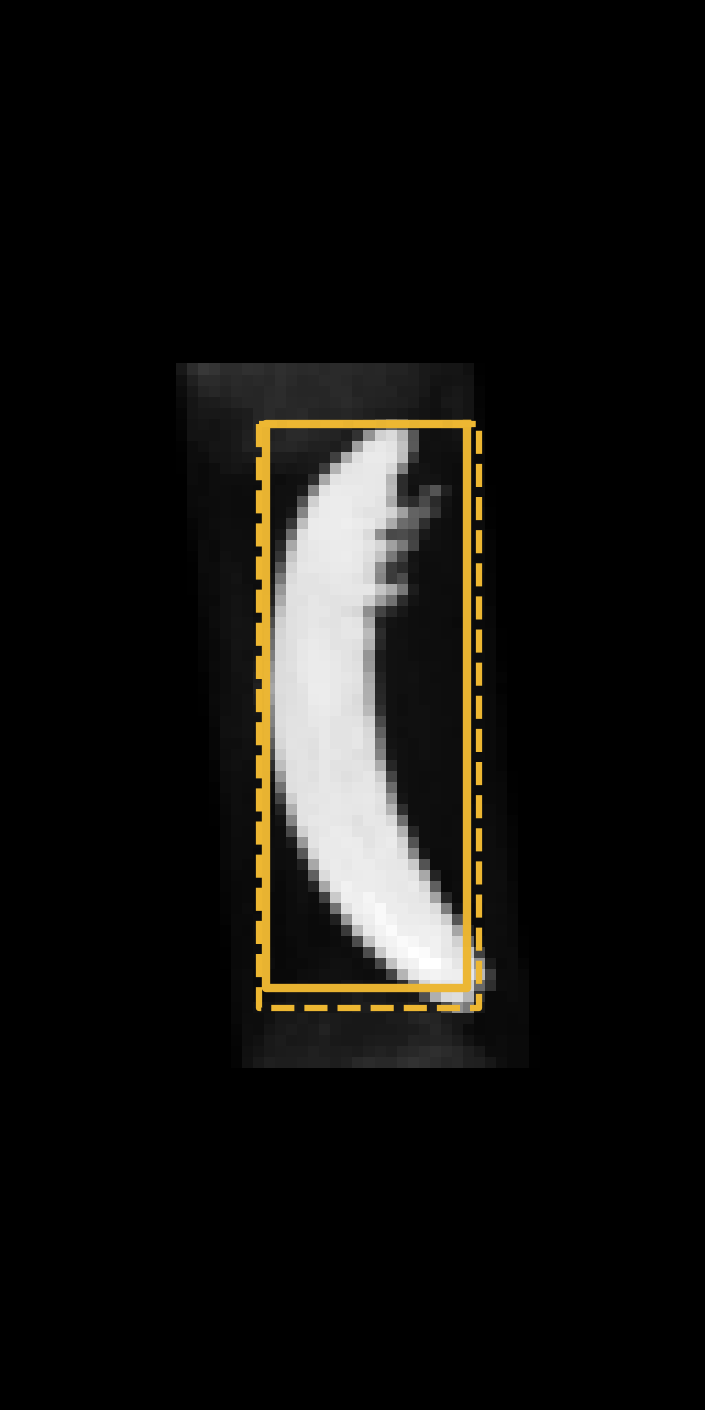}
    \end{tabular} &
    \raisebox{\fix}{\includegraphics[width=0.32\linewidth]{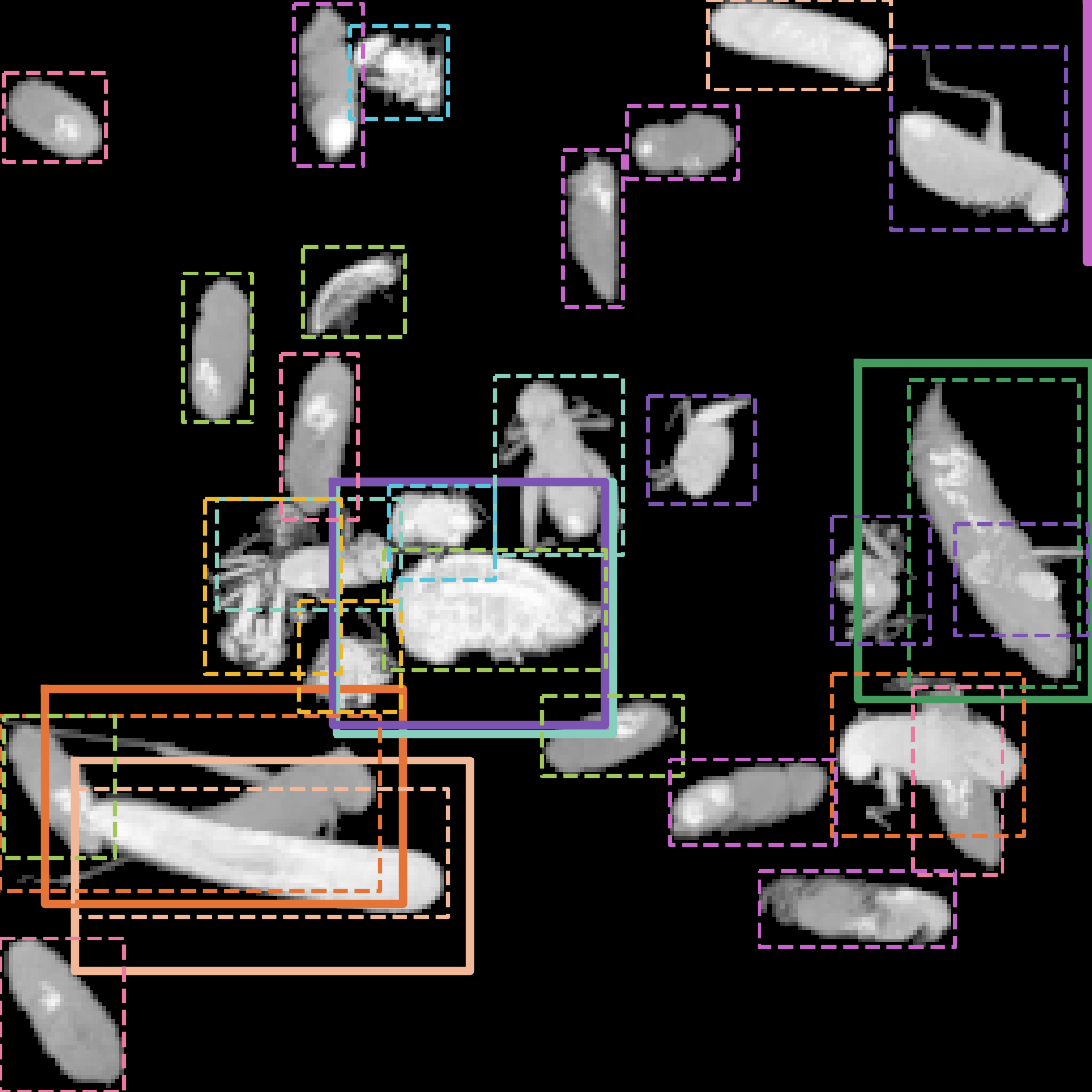}} & 
    \raisebox{\fix}{\includegraphics[width=0.32\linewidth]{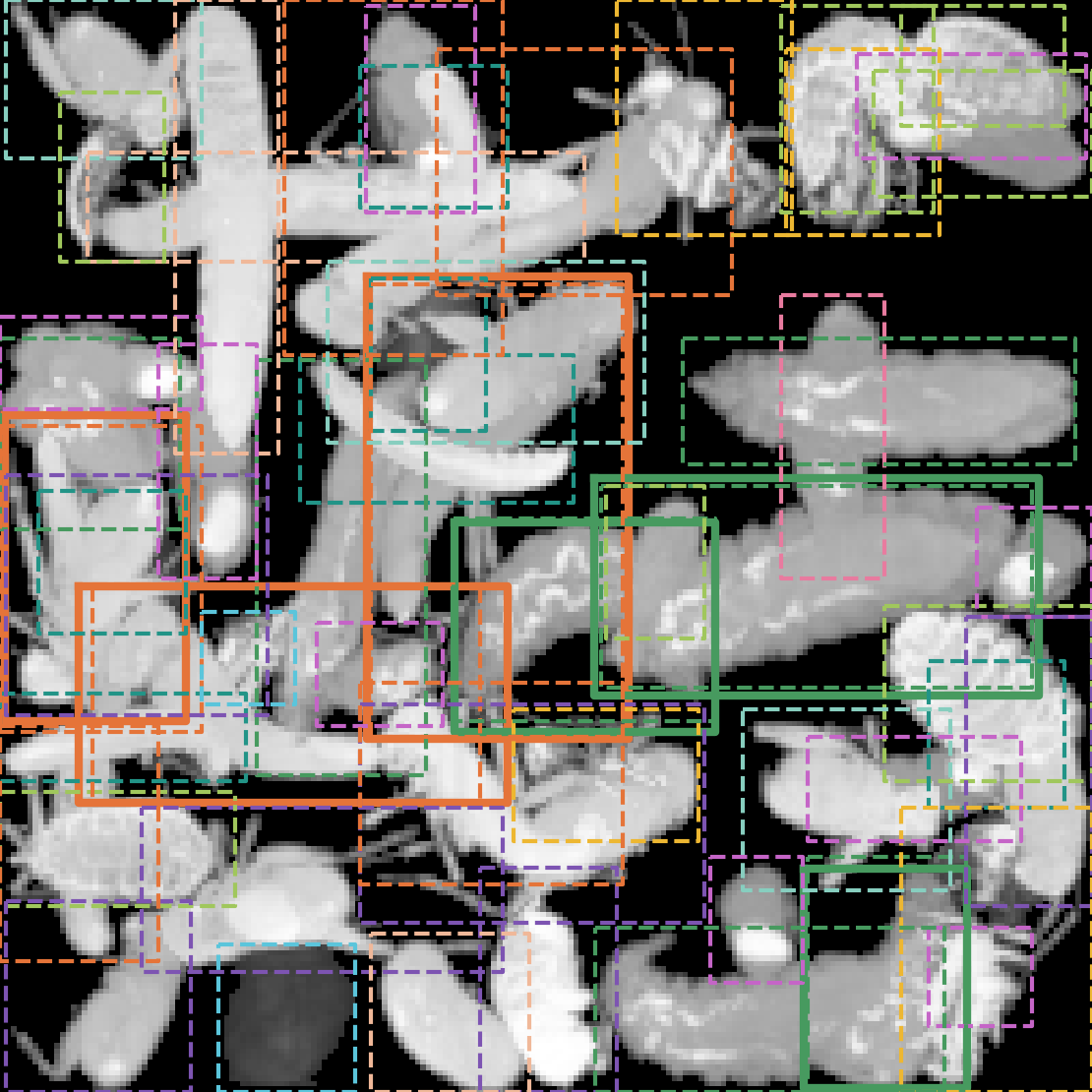}} \\
    \rotatebox[origin=c]{90}{\makebox[\dimexpr 0.32\linewidth + 2pt\relax][c]{nnDetection \cite{baumgartner2021nndetection}}} & 
    \begin{tabular}{cccc}
        \includegraphics[height=\indh]{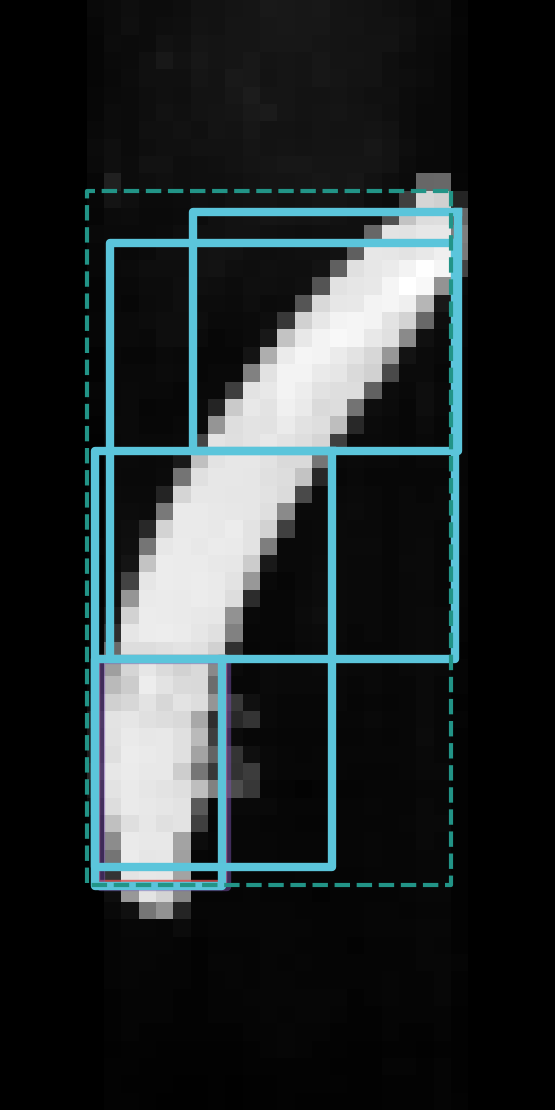} & 
        \includegraphics[height=\indh]{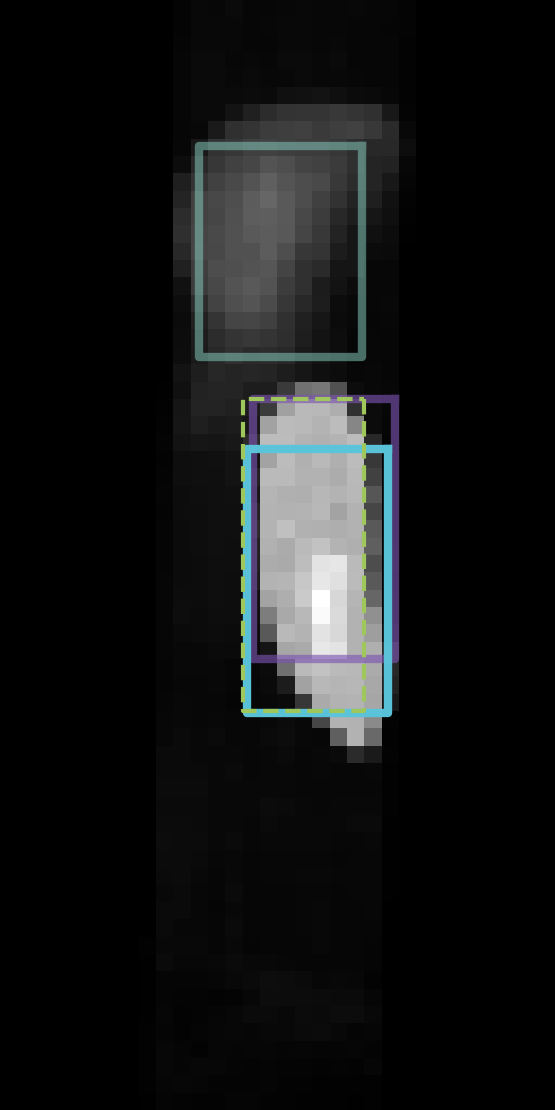} & 
        \includegraphics[height=\indh]{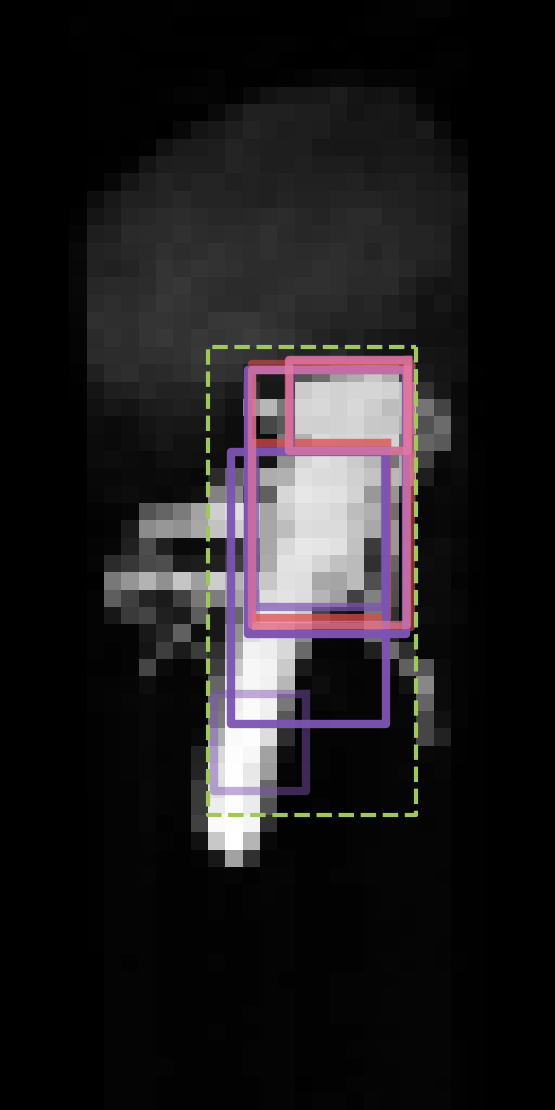} &
        \includegraphics[height=\indh]{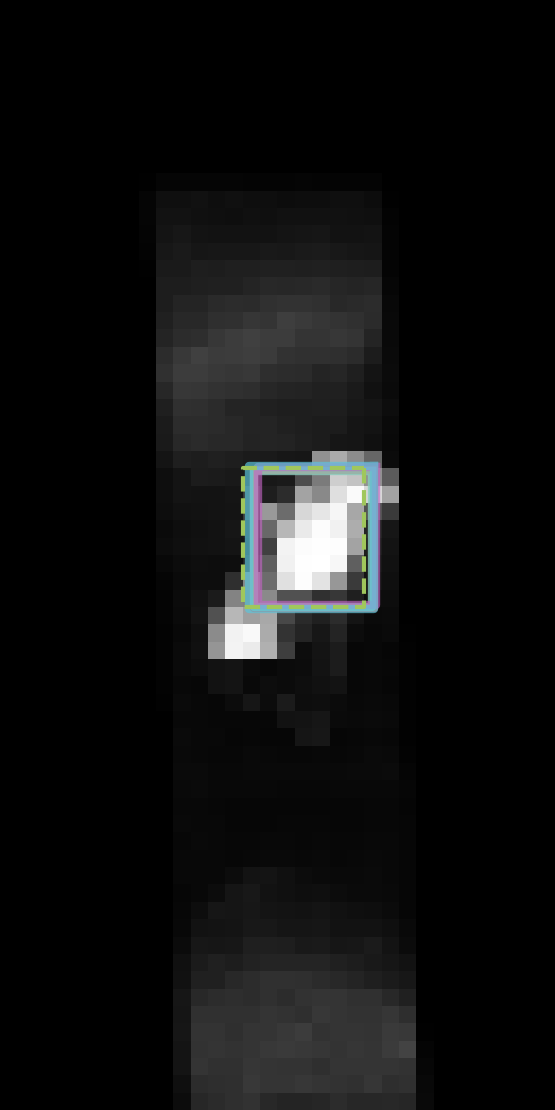} \\
        \includegraphics[height=\indh]{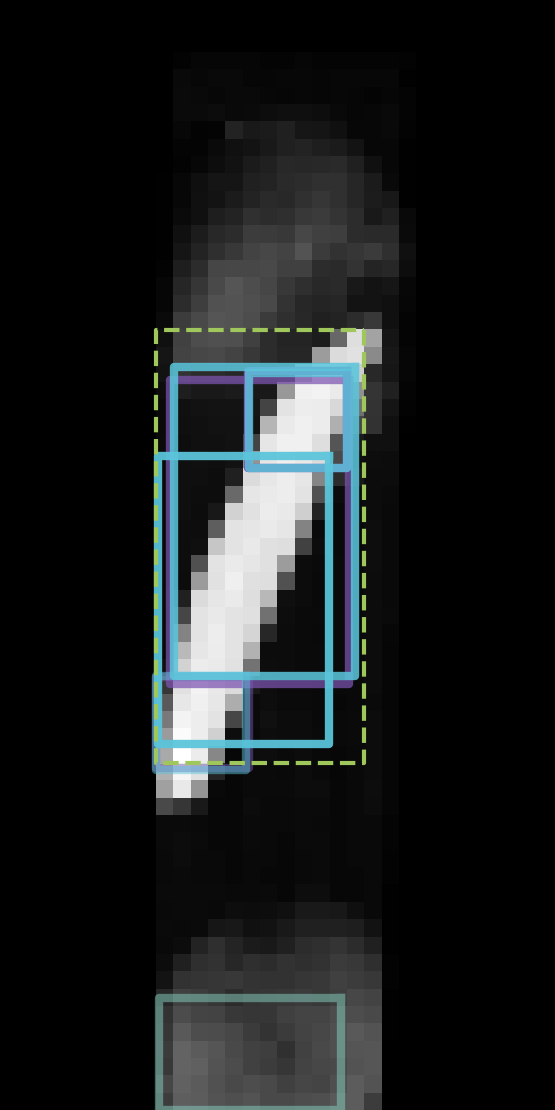} & 
        \includegraphics[height=\indh]{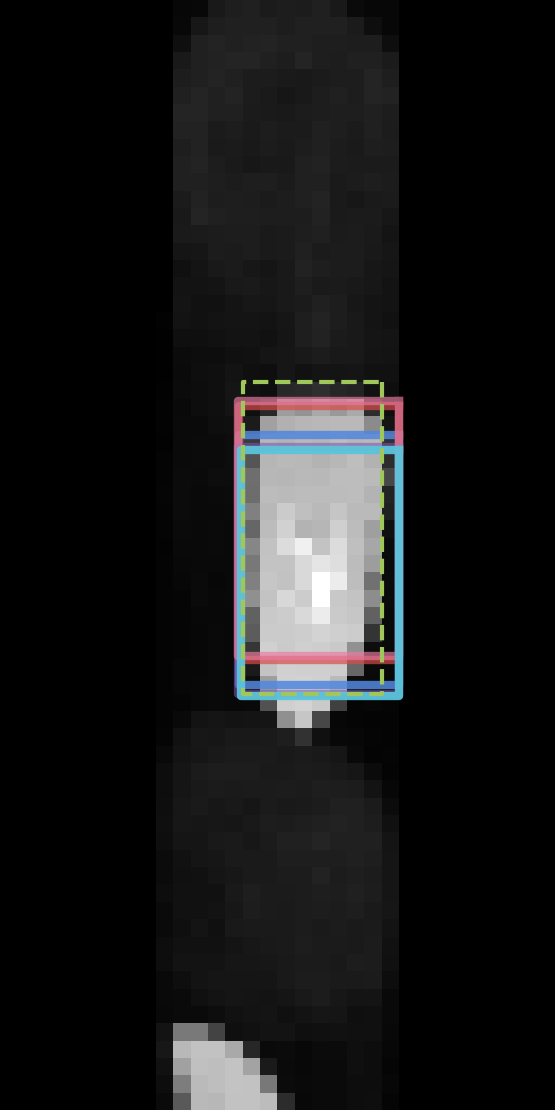} & 
        \includegraphics[height=\indh]{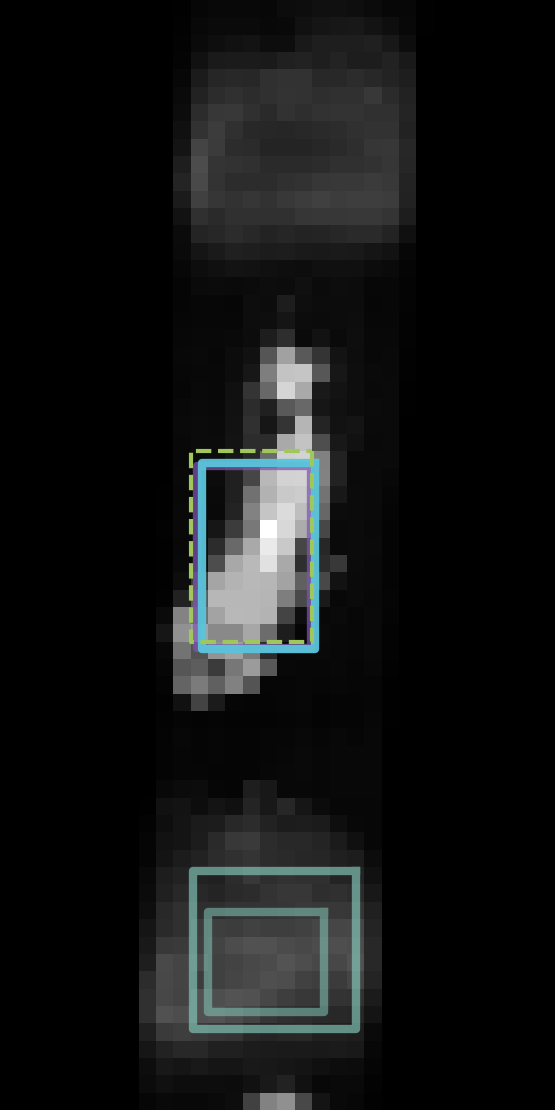} & 
        \includegraphics[height=\indh]{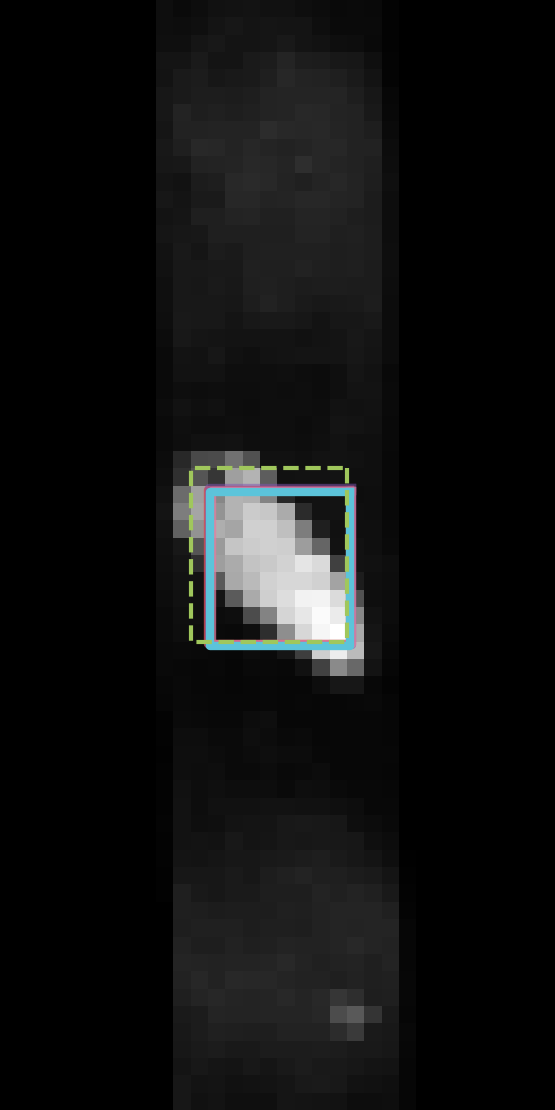}
    \end{tabular} &
    \raisebox{\fix}{\includegraphics[width=0.32\linewidth]{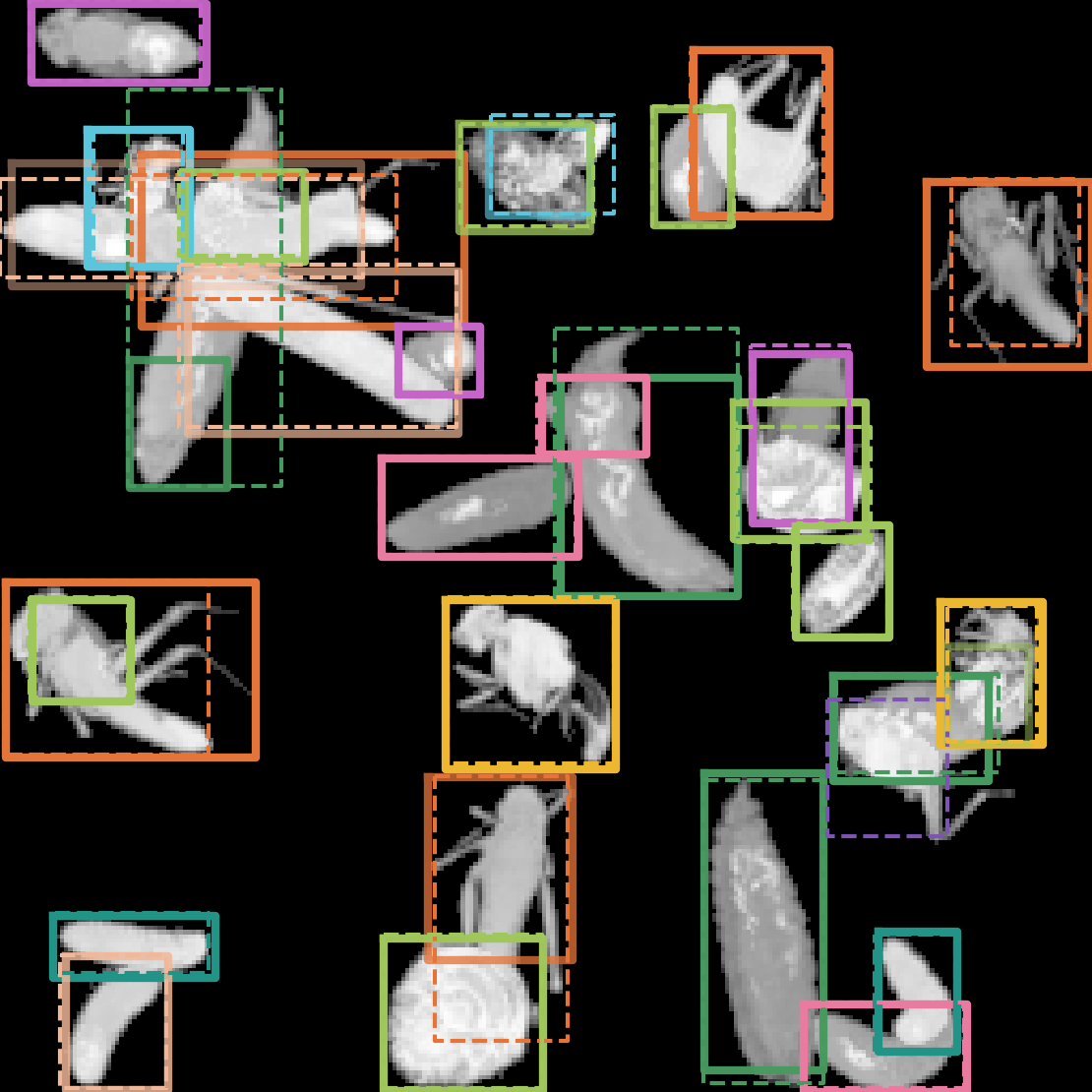}} &
    \raisebox{\fix}{\includegraphics[width=0.32\linewidth]{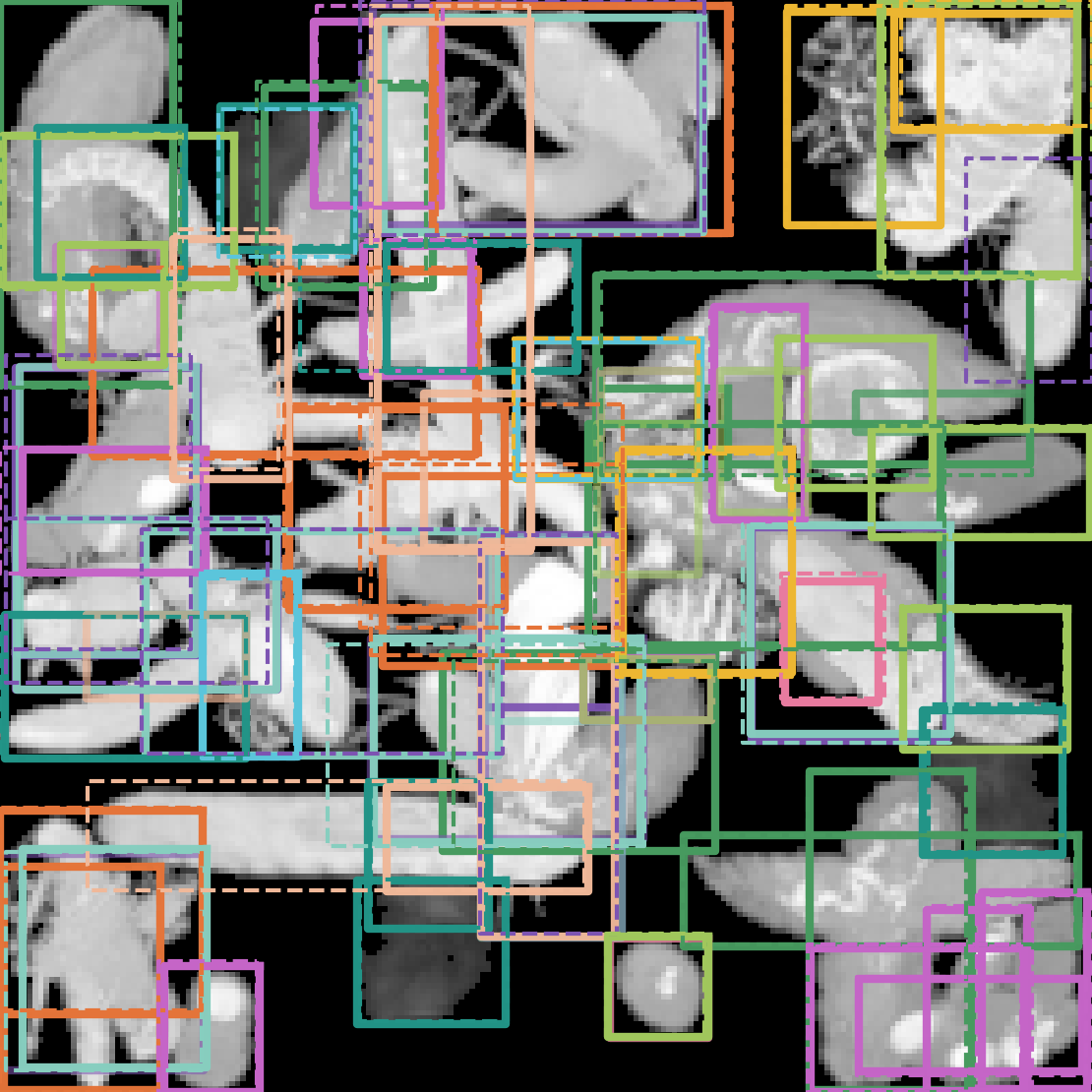}} \\[2.85cm]
    & Individual bugs & Synthetic mixes & Crowded synthetic mixes
    \end{tabular}
    \caption{Examples of detection on the volumes of individual bugs and synthetic mixtures. The ground truth is shown in dashed lines and the detection is shown in solid lines. Colors indicate the class label.}
    \label{fig:mix_detection}
\end{figure*}